\documentclass[11pt]{article}

\UseRawInputEncoding
\usepackage[final]{acl}

\usepackage{times}
\usepackage{latexsym}
\usepackage{pgfplots}
\usepackage{dblfloatfix}
\usepgfplotslibrary{groupplots}

\usepackage[T1]{fontenc}

\usepackage[utf8]{inputenc}

\usepackage{microtype}

\usepackage{inconsolata}

\usepackage{graphicx}

\usepackage{pgfplots}
\usepackage{subcaption}

\usepackage{tabularx}
\usepackage{algorithm}
\usepackage{algpseudocode}
\usepackage{colortbl}   
\usepackage{xcolor}     

\usepackage[dvipsnames]{xcolor} 

\usepackage{amsmath,amsfonts}
\usepackage{graphicx}
\usepackage{textcomp}
\usepackage{xcolor,soul}
\def\BibTeX{{\rm B\kern-.05em{\sc i\kern-.025em b}\kern-.08em
    T\kern-.1667em\lower.7ex\hbox{E}\kern-.125emX}}

\usepackage{booktabs}   
\usepackage{subcaption} 
\usepackage{array}
\usepackage{ragged2e}
\usepackage{float}
\usepackage{listings}
\usepackage{xspace}
\usepackage{multirow}
\usepackage{amsthm}
\usepackage{balance}
\usepackage{tikz}         
\usepackage{adjustbox}    
\usepackage{xcolor}       
\usepackage{listings}

\usepackage[most]{tcolorbox}

\usepackage{xcolor,pifont}
\newcommand*\colourcheck[1]{%
	\expandafter\newcommand\csname #1check\endcsname{\textcolor{#1}{\ding{52}}}%
}

\usepackage{enumitem}
\usepackage[normalem]{ulem} 
\usepackage{array}
\usepackage{wrapfig}
\usepackage{amsmath,amsfonts}

\usepackage{graphicx}
\usepackage{textcomp}
\usepackage{float}
\usepackage{listings}
\usepackage{xspace}
\usepackage{multirow}
\usepackage{amsthm}

\usepackage{balance}
\usepackage{algorithm}
\usepackage{algpseudocode}
\usepackage{colortbl}
\usepackage{subcaption}

\usepackage{xcolor}
\usepackage{multicol}
\usepackage{soul}
\usepackage{framed}
\usepackage{booktabs}
\usepackage{booktabs}
\usepackage{multirow}
\usepackage{siunitx}
\usepackage{multicol}
\usepackage{fancyvrb}
\usepackage{hyperref}
\usepackage{url}

\title{ReVEL: Multi-Turn Reflective LLM-Guided Heuristic Evolution via Structured Performance Feedback}

\author{
  \textbf{Van Duc Cuong}\textsuperscript{1,2},
  \textbf{Nguyen Dinh Tuan Minh}\textsuperscript{2}\thanks{Equal contribution.},
  \textbf{Vu Duc Tam}\textsuperscript{2,*},
  \textbf{Vu Duy Tung}\textsuperscript{4},
  \\
  \textbf{Nguyen Van Son}\textsuperscript{3},
  \textbf{Nguyen Thi Hanh}\textsuperscript{5},
  \textbf{Huynh Thi Thanh Binh}\textsuperscript{2}
  \\
  \\
\textsuperscript{1}VinMotion,
\textsuperscript{2}Hanoi University of Science and Technology,
\textsuperscript{3}Phenikaa School of Computing, \\ Phenikaa  University 
\textsuperscript{4}VinUniversity,
\textsuperscript{5}Phenikaa School of Interdisciplinary Digital \\ Technology, Phenikaa University.
\\
\small{
    \textsuperscript{*}\textbf{Correspondence:} \href{mailto:hanh.nguyenthi@phenikaa-uni.edu.vn}{\texttt{hanh.nguyenthi@phenikaa-uni.edu.vn}}
  }
}

\begin{document}
\lstset{literate={–}{-}{1}}

\maketitle

\begin{figure*}[t]
    \centering
    \includegraphics[width=\textwidth]{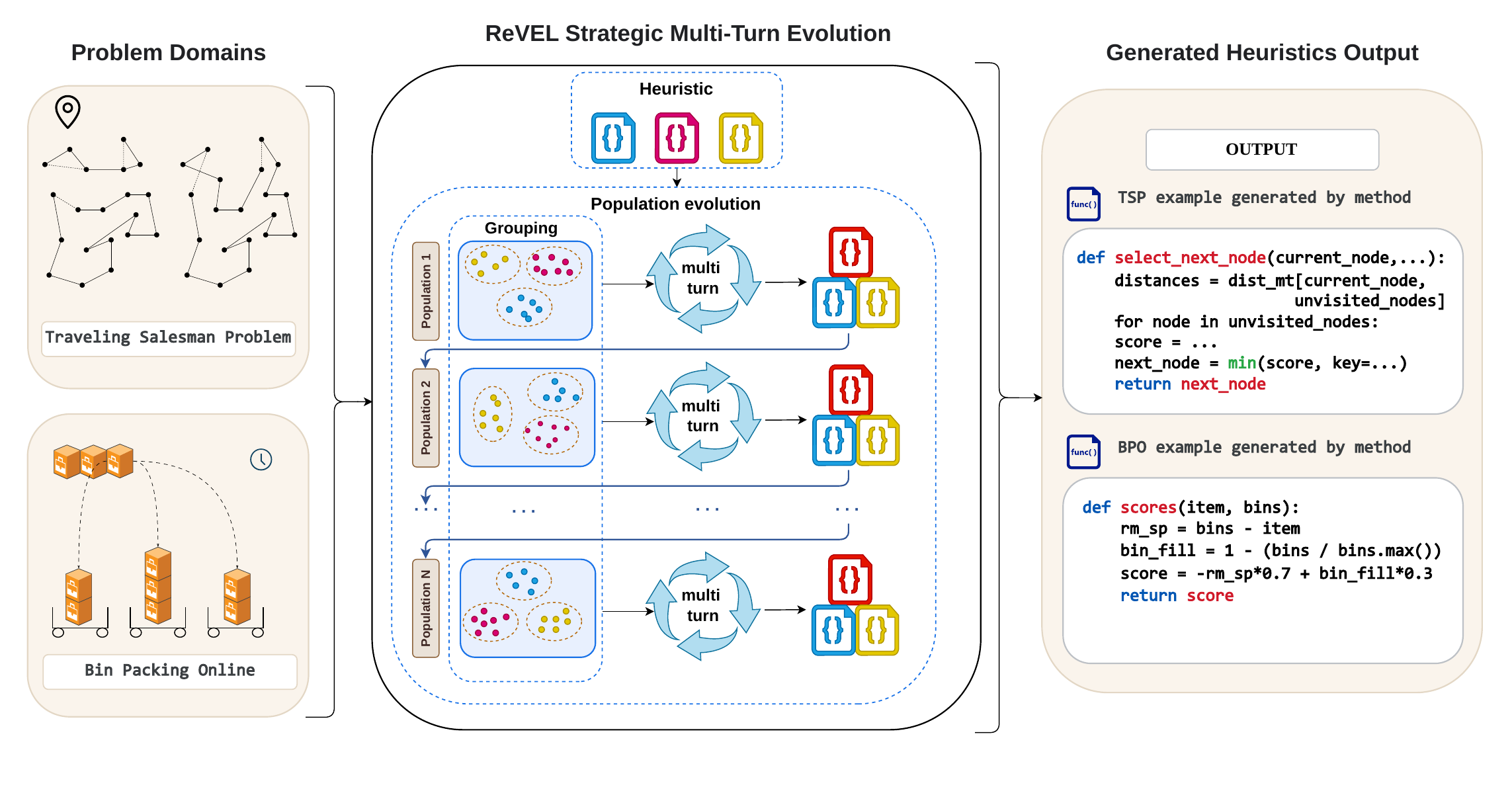}
    \vspace{-0.9cm}
\caption{
The proposed \textbf{\textsc{REVEL}} framework addresses combinatorial optimization problems across diverse domains and problem scales through strategic multi-turn heuristic evolution. Starting from initial candidate search strategies, \textsc{REVEL} organizes heuristic populations into collaborative refinement groups that iteratively perform evaluation, strategic rewriting, and evolutionary adaptation. Through repeated multi-turn interactions, the framework progressively explores and refines increasingly effective search behaviors, enabling the automatic discovery of high-performing heuristics for complex combinatorial optimization problems.
}
\label{fig:overview}
\end{figure*}

\begin{abstract}
Designing effective heuristics for NP-hard combinatorial optimization problems remains challenging and often requires substantial domain expertise. Recent LLM-guided evolutionary methods have shown promise for automated heuristic generation, but most existing approaches refine heuristics independently or through limited pairwise feedback. We propose \textbf{ReVEL}: Multi-Turn \textbf{\underline{Re}}flecti\textbf{\underline{v}}e LLM-Guided Heuristic \textbf{\underline{E}}vo\textbf{\underline{l}}ution via Structured Performance Feedback, a framework for group-wise multi-turn heuristic refinement. ReVEL organizes heuristics into behavior-aware reflective groups, including similarity-driven groups for localized refinement and diversity-driven groups for exploratory search. Within each group, the LLM performs iterative multi-turn refinement using accumulated performance feedback, enabling related heuristics to be jointly analyzed and progressively improved across evolutionary iterations. Experiments on standard combinatorial optimization benchmarks show that ReVEL generally improves optimization performance over existing LLM-guided evolutionary baselines across multiple settings and LLM backbones. Additional analyses suggest that behavior-aware grouping contributes to more consistent refinement trajectories during iterative heuristic evolution.
\end{abstract}

\section{Introduction}
\label{sec:intro}

Large language models (LLMs) have recently demonstrated strong capabilities in program synthesis, iterative refinement, and feedback-driven code generation~\cite{liao2026explainable, gozeten2026continuous}. These advances have motivated growing interest in using LLMs for combinatorial optimization, where heuristics can be progressively improved through repeated interaction with optimization feedback.

Combinatorial optimization problems (COPs), such as the Traveling Salesman Problem (TSP)~\cite{bock2025survey} and Bin Packing Problem (BPP)~\cite{aider2025adaptative}, require searching exponentially large discrete spaces in which high-quality solutions typically emerge through sustained iterative refinement.

\underline{First}, \textbf{classical heuristic and meta-heuristic methods}~\cite{almufti2025comparative, rondano2025heuristic} rely heavily on manually designed search operators tailored to problem-specific structures. While effective, these methods depend strongly on expert knowledge and fixed optimization strategies.

\underline{Second}, \textbf{hyper-heuristic methods}~\cite{xu2025adaptive, vela2025hyper} attempt to automate heuristic selection and composition. However, optimization remains constrained by predefined heuristic components, limiting the discovery of fundamentally new search behaviors.

\underline{Third}, recent \textbf{LLM-guided evolutionary frameworks} such as EoH~\cite{liu2024evolution} and ReEvo~\cite{ye2024reevo} demonstrate that LLMs can directly generate and iteratively refine optimization heuristics using evolutionary feedback. Despite promising results, existing methods primarily rely on reflection over isolated candidates or pairwise comparisons. Consequently, optimization feedback remains fragmented across generations, limiting the model's ability to identify broader behavioral patterns within the evolving heuristic population.

These limitations suggest that the effectiveness of reflective heuristic optimization depends critically on how optimization feedback is organized and exposed to the model during refinement.

\textbf{Our approach:} We propose \textsc{ReVEL}, a reflective evolutionary framework that iteratively refines behavior-aware heuristic groups. Rather than refining heuristics independently, \textsc{ReVEL} organizes heuristics according to shared optimization behavior, enabling the LLM to analyze recurring strengths, common failure modes, and complementary search patterns within each group.

Based on these grouped reflective contexts, \textsc{ReVEL} performs iterative refinement that balances exploratory heuristic modification and exploitative local improvement throughout optimization. This grouped-refinement process produces more coherent, reflective feedback, leading to more stable optimization trajectories and improved long-horizon search performance.

Extensive experiments across diverse combinatorial optimization problems demonstrate that \textsc{ReVEL} achieves substantial improvements over strong LLM-guided evolutionary baselines across problem settings and LLM backbones. Further analysis reveals that behavior-aware grouping enables more reliable refinement and stable optimization throughout evolutionary search.

\textbf{Contributions.}
Our contributions are summarized as follows:

\begin{itemize}

\item \textbf{Reflective evolutionary optimization.}
We introduce \textsc{ReVEL}, a reflective evolutionary framework that reformulates LLM-guided heuristic optimization as iterative refinement over grouped heuristic populations.

\item \textbf{Behavior-aware reflective grouping.}
We propose a grouping mechanism that organizes heuristics according to shared optimization behavior, enabling more coherent reflective feedback during heuristic refinement.

\item \textbf{Iterative grouped refinement.}
We develop a multi-turn refinement strategy that balances exploratory heuristic modification and exploitative local improvement through grouped reflective optimization.

\item \textbf{Comprehensive empirical analysis.}
We demonstrate across multiple combinatorial optimization benchmarks that structured reflective grouping substantially improves optimization quality, refinement consistency, and robustness across different LLM backbones.

\end{itemize}
\section{Related work}

\subsection{Automatic Heuristic Design}
Traditional heuristic design relies on manually crafted rules tailored to problem-specific structures, limiting scalability and cross-domain transfer~\cite {Burke}. Genetic Programming (GP) and hyper-heuristics partially automate this process by evolving heuristics from predefined operators or heuristic pools~\cite{Branke2016}. More recently, Neural Combinatorial Optimization (NCO) methods use Deep Reinforcement Learning to learn constructive policies directly from problem instances~\cite{bello2017neuralcombinatorialoptimizationreinforcement, Kool2018AttentionLT}. While these approaches reduce manual design effort, they often require substantial training and remain sensitive to representations, reward design, or predefined search structures. More importantly, they do not explicitly support iterative language-based reasoning over evolving heuristic behavior and execution feedback.

\subsection{LLMs for Evolutionary Search}
Large Language Models (LLMs)~\cite{NIPS2017_3f5ee243,zhao2023survey} provide a flexible alternative by generating and modifying executable programs through natural-language and code-level feedback. This has enabled LLMs to act as semantic variation operators within evolutionary optimization frameworks~\cite{chi2026generalized,huang2026heuristic}. In contrast to large-scale sampling and filtering approaches such as AlphaCode~\cite{Li_2022}, recent methods increasingly couple program generation with evaluator-guided evolutionary search.
FunSearch~\cite{RomeraParedes2024} pioneered this direction by integrating LLM program generation with iterative evaluation and selection. EoH~\cite{liu2024evolution} further co-evolves heuristic programs and natural-language descriptions, while LLaMEA~\cite{LLaMEA} uses LLMs as semantic mutation and crossover operators. These methods substantially improve automated heuristic discovery, but refinement remains largely driven by individual candidates or local comparisons, limiting the population-level structure the model sees during evolution.

\subsection{Multi-Turn Reflection}
Multi-turn reflection provides a complementary mechanism for progressively improving LLM outputs through feedback. Self-Refine~\cite{madaan2023selfrefine} and Reflexion~\cite{shinn2023reflexion} demonstrates iterative self-correction through accumulated feedback, while Self-Debugging~\cite{chen2024teaching} incorporates execution signals such as unit tests and runtime traces to refine generated programs.
Within heuristic optimization, ReEvo~\cite{ye2024reevo} introduces reflective evolution by generating verbal critiques from pairwise performance differences and using them to refine heuristic code. HSEvo~\cite{dat2025hsevo} promotes population diversity during evolutionary search. However, existing reflective approaches primarily reason over isolated heuristics or pairwise relations, leaving population-wide behavioral structure and long-horizon refinement history underutilized. \textsc{ReVEL} addresses this gap by organizing heuristics into behavior-aware reflective groups and performing multi-turn refinement over accumulated population-level feedback. This shifts heuristic evolution from repeated local refinement toward structured reasoning over an evolving heuristic population.
\section{Methodology}
\label{sec:method}
\subsection{Overview}

\begin{figure*}[ht]
    \centering
    \includegraphics[width=1.0\textwidth]{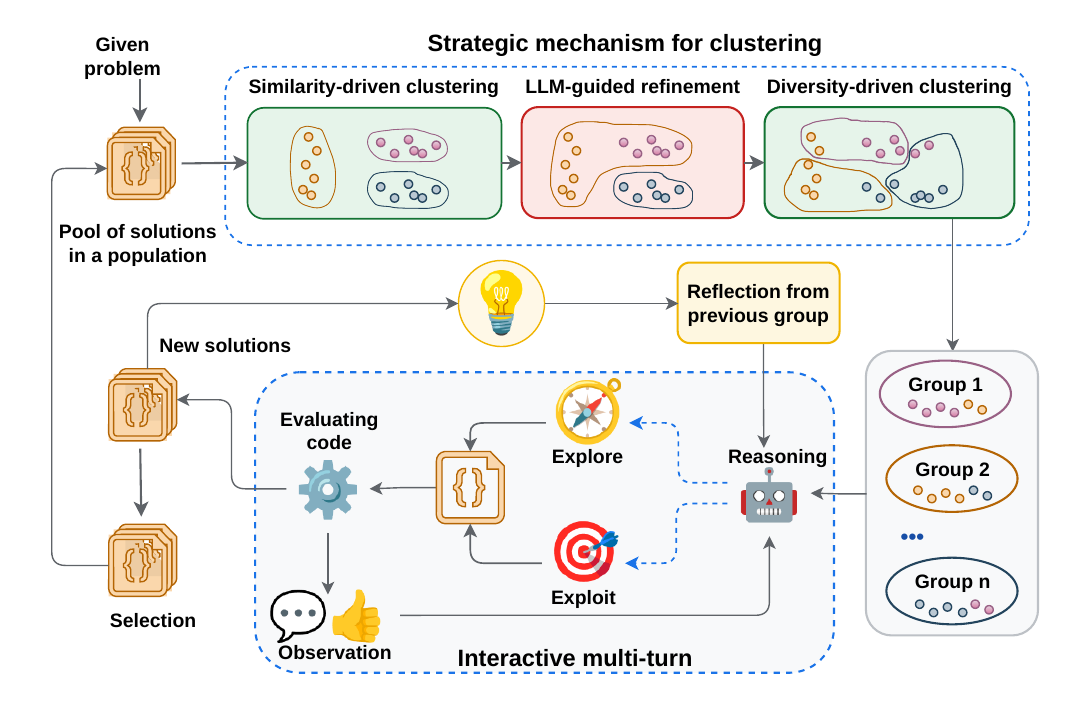}
\vspace{-0.8cm}
\caption{\textbf{\textsc{ReVEL}: Multi-turn reflective evolutionary optimization through structured population feedback.} Starting from an initial heuristic population, \textsc{ReVEL} iteratively organizes heuristics into behaviorally coherent groups, performs reflective refinement over accumulated population feedback, and evolves improved heuristics through population-level reasoning.}
\label{fig:pipeline}
\vspace{-0.3cm}
\end{figure*}

Figure~\ref{fig:pipeline} illustrates the architecture of \textsc{ReVEL}, a reflective evolutionary framework that integrates LLMs into the optimization loop as population-level reflective refinement framework. Unlike existing LLM-guided evolutionary frameworks that refine heuristics through isolated reflection, \textsc{ReVEL} maintains persistent reflective reasoning across optimization iterations as populations evolve.

\textbf{At a high level,} \textsc{ReVEL} formulates heuristic evolution as an iterative refinement process guided by accumulated population-level feedback. Starting from an initial heuristic population, the framework repeatedly evaluates collective search behavior, organizes heuristics into behaviorally coherent groups, and performs reflective refinement based on population-level optimization trends rather than isolated candidate performance.

Central to \textsc{ReVEL} is a grouped reflective refinement mechanism that balances behavioral coherence with search diversity. By analyzing grouped optimization patterns, the framework reveals recurring strengths, shared failures, and complementary search behaviors that are difficult to infer from isolated heuristic comparisons. Guided by these reflective signals, the LLM iteratively refines existing heuristics or synthesizes new strategies through optimization feedback. Reflective feedback accumulates across generations, enabling progressively more informed refinement. Through this interaction between grouped reflective refinement and iterative optimization feedback, \textsc{ReVEL} reformulates heuristic evolution from independent candidate generation into group-aware refinement.

\vspace{-0.3cm}

\subsection{Behavior-Aware Heuristic Grouping}
\label{section:grouping}
\paragraph{Representation.}

Each heuristic $h$ is evaluated on a benchmark set $\mathcal{I}=\{1,\dots,m\}$ with objective values $e_i(h)$ on instance $i$. To characterize optimization behavior across instances, we represent each heuristic using a normalized performance-profile vector: $\mathbf{z}(h)
    =
    \frac{\mathbf{e}(h)-\mathbf{e}^{\ast}}
    {\mathbf{e}^{\ast}},$ where
$
\mathbf{e}(h)=\big(e_1(h),\dots,e_m(h)\big)^\top
$
denotes the performance vector of heuristic $h$, and
$
\mathbf{e}^{\ast}
=
\big(e_1^{\ast},\dots,e_m^{\ast}\big)^\top
$
contains the best-known objective values within the current heuristic population: $e_i^{\ast}
    =
    \min_{h' \in \mathcal{H}} e_i(h'), i=1,\dots,m.$
This representation captures the relative behavioral profile of each heuristic and provides a normalized space for population-level comparison.

\paragraph{Structured reflective grouping.}

Reflective optimization requires balancing two competing objectives: coherent diagnosis and diverse exploration. Similar heuristics reveal recurring optimization patterns for precise refinement, whereas diverse heuristics expose complementary search behaviors beyond local optima. \textsc{ReVEL} resolves this trade-off through homogeneous groups for coherent reflection and heterogeneous groups for exploration.

\subsubsection*{Homogeneous groups (similarity-driven)}

To construct behaviorally coherent groups, we measure similarity between heuristics using both optimization behavior and program structure. Behavioral similarity is computed in the normalized performance space using cosine similarity:
\vspace{-0.3cm}
\begin{equation}
    \mathrm{sim}_{\mathrm{perf}}(h_i,h_j)
    =
    \frac{
    \mathbf{z}(h_i)^\top \mathbf{z}(h_j)
    }{
    \|\mathbf{z}(h_i)\|_2
    \|\mathbf{z}(h_j)\|_2
    }.
\label{equa1}
\end{equation}

\vspace{-0.6cm}
\begin{equation}
    \mathrm{sim}_{\mathrm{code}}(h_i,h_j)
    =
    \mathrm{CodeBLEU}(h_i,h_j).
    \label{equa2}
\end{equation}
(\ref{equa1}) captures whether heuristics exhibit similar optimization dynamics across benchmark instances. (\ref{equa2}) measures structural similarity between heuristics using CodeBLEU, which incorporates lexical, syntactic, semantic, and data-flow information. The resulting similarity score is defined as:
\begin{equation}
    \mathrm{sim}(h_i,h_j)
    =
    \alpha \,
    \mathrm{sim}_{\mathrm{perf}}(h_i,h_j)
    +
    \beta \,
    \mathrm{sim}_{\mathrm{code}}(h_i,h_j),
\end{equation}

where $\alpha,\beta \geq 0$ balance behavioral and structural similarity. Using these scores, we represent the heuristic population as a weighted similarity graph $G=(H,E,W)$, 
    $W_{ij}=\mathrm{sim}(h_i,h_j)$, where $H$ denotes the heuristic set and $W$ stores pairwise similarities. The similarity matrix is normalized and transformed into a dissimilarity matrix:
\vspace{-0.1cm}
\begin{equation}
    \tilde{W}
    =
    \frac{W}{\max_{p,q}W_{pq}},
    \qquad
    d_{ij}
    =
    1-\tilde{W}_{ij}.
\end{equation}

Because reflective quality is highly sensitive to the granularity of grouping, \textsc{ReVEL} adopts a two-stage clustering strategy. First, agglomerative clustering with linkage function $L$ produces an intentionally fine-grained over-partition:
\vspace{-0.3cm}
\begin{equation}
    (C_1,\dots,C_m)
    =
    \operatorname{Agglomerative}_{L}(D;m),
\end{equation}

where $m \gg 1$ is selected such that each cluster remains locally coherent: 
\begin{equation}
    \mathrm{diam}(C_\ell)
    =
    \max_{x,y \in C_\ell} d(x,y)
    \leq \delta,
    \qquad
    \delta \ll 1.
\end{equation}

This over-partition preserves localized behavioral structure while avoiding premature semantic merging. The resulting partition
$
\mathcal{C}_0=\{C_1,\dots,C_m\}
$
is subsequently refined by the LLM:

\begin{equation}
    \mathcal{C}^{\mathrm{ref}}
    =
    \Phi_{\mathrm{LLM}}(\mathcal{C}_0,W),
\end{equation}

\hspace{-0.4cm}subject to $    \bigcup_{C \in \mathcal{C}^{\mathrm{ref}}} C = H,
    C_p \cap C_q = \emptyset
    \quad (p \neq q).$
    
This refinement stage allows the LLM to globally reorganize clusters according to higher-level semantic and behavioral relationships that static similarity metrics alone may fail to capture.

\subsubsection*{Heterogeneous groups (diversity-driven)}

While homogeneous groups support precise reflective analysis, excessive similarity can narrow exploration and encourage convergence toward locally dominant strategies. To broaden the reflective search space, \textsc{ReVEL} additionally constructs heterogeneous groups by combining heuristics from distinct homogeneous clusters. For a homogeneous cluster
$
G=\{h_1,\dots,h_m\},
$
We estimate internal diversity using the entropy of pairwise-similarity distributions. Defining normalized affinities: 
\begin{equation}
    p_{ij}
    =
    \frac{
    \mathrm{sim}(h_i,h_j)
    }{
    \sum_{u<v}
    \mathrm{sim}(h_u,h_v)
    },
    \qquad
    i<j,
\end{equation}

The cluster entropy is computed as: $\mathcal{H}(G)
    =
    -
    \sum_{i<j}
    p_{ij}\log p_{ij}.$ Higher entropy indicates greater structural and behavioral diversity within the cluster. Given homogeneous clusters
$
\{G_1,\dots,G_k\},
$
Sampling weights are assigned proportionally to entropy: 
\begin{equation}
    w_i
    =
    \frac{
    \mathcal{H}(G_i)
    }{
    \sum_{j=1}^{k}
    \mathcal{H}(G_j)
    }.
\end{equation}

For a target heterogeneous group size $L$, each cluster contributes $L_i= \lfloor w_i L \rfloor$ heuristics sampled uniformly from $G_i$. The final heterogeneous group is $G^{\mathrm{het}}
    =
    \bigcup_{i=1}^{k} S_i,
    S_i \subseteq G_i,
    |S_i| = L_i.$

By exposing the LLM to structurally diverse optimization behaviors, heterogeneous grouping expands the reflective context available during refinement and encourages exploration beyond locally coherent heuristic families.

\begin{table*}[t]
\centering
\caption{Analysis of Reasoning Categories, their frequency, and examples.}
\label{tab:reasoning_breakdown}

\small
\setlength{\tabcolsep}{5pt}
\renewcommand{\arraystretch}{1.05}

\begin{tabularx}{\textwidth}{@{}>{\raggedright\arraybackslash}p{0.22\textwidth} >{\raggedright\arraybackslash}X c >{\raggedright\arraybackslash\scriptsize}X@{}}
\toprule
\textbf{Reasoning Category} & \textbf{Description} & \textbf{Freq} & \textbf{Example} \\
\midrule

\textit{Structure rewrite} 
& Completely changes the algorithm family. 
& 16 
& Given that the highest-performing known approach is ``Stabilized Harmonic-Arctanh'' $\ldots$ I should explore this proven algorithmic family rather than tuning simpler approaches. \\

\textit{Heuristic modification} 
& Same algorithm family and pipeline structure, but decision and scoring logic is substantially rewritten. 
& 89 
& Given the significant performance gap and clear indication that Worst-Fit works better, we should focus on refining this approach $\ldots$ \\

\textit{Parameter tuning} 
& Only numeric changes (weights, thresholds, constants) with formulas and pipeline unchanged. 
& 75 
& Given the regression, we should revert to the simpler version and make minimal adjustments to core parameters $\ldots$ \\

\bottomrule
\end{tabularx}
\label{tab:reasoning_category}
\end{table*}

\begin{algorithm}[H]
\caption{\textsc{ReVEL}: Structured Multi-Turn Reflective Evolution}
\label{alg:revel}
\begin{algorithmic}[1]

\Require Instance set $\mathcal{D}$, population size $N$, generations $G$, reflection turns $T$
\Ensure Best heuristic $h^{*}$

\State Initialize population $\mathcal{H}^{(0)}$ with the LLM
\State Initialize search memory $\mathcal{M}_0 \gets \emptyset$

\For{$g=1,\ldots,G$}

    \State Evaluate $\mathcal{H}^{(g-1)}$ and compute behavioral profiles $\mathbf{z}(h)$

    \State Build behavior-aware reflective groups:
    \[
    \mathcal{G}
    \gets
    \textsc{Group}\!\left(
    \mathcal{H}^{(g-1)},
    s_{\mathrm{perf}},
    s_{\mathrm{code}}
    \right)
    \]

    \State $\widetilde{\mathcal{H}}_g \gets \emptyset$

    \For{each group $\mathcal{G}_k \in \mathcal{G}$}

        \State Initialize group feedback $\mathcal{F}_k^{(1)}$

        \For{$t=1,\ldots,T$}

            \State $\rho_k^{(t)}
            \gets
            \textsc{Reflect}
            (\mathcal{G}_k,\mathcal{F}_k^{(t)},\mathcal{M}_{g-1})$

            \State $a_k^{(t)}
            \in
            \{\textsc{Explore},\textsc{Exploit}\}
            \gets
            \textsc{Decide}(\rho_k^{(t)})$

            \State $\tilde{h}_k^{(t)}
            \gets
            \textsc{Generate}_{\mathrm{LLM}}
            (\rho_k^{(t)},a_k^{(t)})$

            \State Evaluate $\tilde{h}_k^{(t)}$ and obtain observation $o_k^{(t)}$

            \State $\mathcal{F}_k^{(t+1)}
            \gets
            \mathcal{F}_k^{(t)}
            \cup
            \{o_k^{(t)}\}$

            \State $\widetilde{\mathcal{H}}_g
            \gets
            \widetilde{\mathcal{H}}_g
            \cup
            \{\tilde{h}_k^{(t)}\}$

        \EndFor
    \EndFor

    \State Update search memory $\mathcal{M}_g$

    \State Evolve population:
    \[
    \mathcal{H}^{(g)}
    \gets
    \operatorname{TopN}
    \left(
    \mathcal{H}^{(g-1)}
    \cup
    \widetilde{\mathcal{H}}_g
    \right)
    \]

\EndFor

\State \Return
$h^{*}=\arg\min_{h\in\mathcal{H}^{(G)}} f(h)$

\end{algorithmic}
\end{algorithm}

\subsection{Multi-turn Reflective Refinement}

Given the groups constructed in Section~\ref{section:grouping}, \textsc{ReVEL} performs iterative multi-turn refinement over grouped heuristics. Refinement operates locally within each group, while selection is performed globally across the population, enabling localized heuristic adaptation without sacrificing population-level diversity.

\paragraph{Group-aware multi-turn refinement.}

At iteration $t$, each reflective group $G_k$ contains heuristics with related optimization behavior. The LLM receives the grouped heuristics together with structured optimization feedback, including relative performance, recurring failure patterns, prior refinement history, and summarized search observations.

Conditioned on the accumulated search memory $M_t$, the LLM performs iterative refinement within each group across multiple turns: $\tilde{\mathcal{H}}_k^{(r+1)}
\sim
\mathcal{G}\left(
G_k,
\tilde{\mathcal{H}}_k^{(r)},
M_t
\right),$ where $r$ denotes the refinement turn and $\mathcal{G}$ denotes the refinement operator. Unlike single-step refinement, the multi-turn process allows the model to revise heuristics using accumulated group-level feedback across turns.

\paragraph{Exploration and exploitation.}

Rather than prescribing exploration through fixed schedules or heuristic rules, \textsc{ReVEL} formulates exploration--exploitation as a reasoning problem. At each refinement turn, the LLM jointly analyzes the current heuristic, accumulated optimization feedback, recurring failure patterns, and the complete multi-turn refinement trajectory to determine the most promising refinement strategy.

Instead of blindly generating another candidate, the LLM explicitly decides whether optimization would benefit more from \emph{exploration}, by proposing structurally different heuristic families, or from \emph{exploitation}, by refining the existing heuristic through localized algorithmic modifications and parameter adaptation. Because this decision is conditioned on the evolving optimization trajectory rather than a predefined schedule, refinement naturally adapts to the search state, remaining exploratory under stagnation while progressively focusing on promising heuristic families as evidence accumulates.

\paragraph{Search memory.}
To maintain refinement continuity across iterations, \textsc{ReVEL} maintains compressed search memory that accumulates grouped optimization feedback across evolutionary iterations. The memory captures recurring successes, failures, explored search directions, and refinement outcomes throughout the optimization process.

\paragraph{Population update.}

Refined heuristics generated across all groups are evaluated and merged with the current population. The next generation is formed using elitist selection: $\mathcal{H}_{t+1}
=
\mathrm{TopK}
(
\mathcal{H}_t
\cup
\tilde{\mathcal{H}}_t
),$ where $\tilde{\mathcal{H}}_t$ denotes all heuristics generated through group-aware multi-turn refinement. By combining intra-group refinement with global evolutionary selection, \textsc{ReVEL} maintains diverse search trajectories while progressively improving promising heuristic families across optimization iterations.
\section{Experiments}
\label{headings}

\subsection{Experimental Setup}

We evaluate \textsc{ReVEL} through $3$ research questions:

\textbf{RQ1. [Effectiveness]} 
How effectively does \textsc{REVEL} improve heuristic optimization over existing evolutionary and LLM-guided methods?


\textbf{RQ2. [Heuristic Evolution]}
How do behavior-aware grouping and multi-turn refinement shape heuristic evolution during optimization?

\textbf{RQ3. [Ablation Study]} 
How does each component of \textsc{ReVEL} contribute to overall performance?

\textbf{RQ4. [Cost Efficiency]} How cost-efficient is REVEL in improving optimization quality?

\paragraph{Benchmarks and baselines.}

We evaluate \textsc{ReVEL} on two representative combinatorial optimization benchmarks: the Traveling Salesman Problem (TSP) and online Bin Packing Problem (BPP). Additional results on CVRP and TSPLib benchmarks are provided in the Appendix.

We compare against recent LLM-based optimization frameworks, including EoH~\cite{liu2024evolution} and ReEvo~\cite{ye2024reevo}, as well as classical heuristic baselines.

\paragraph{Implementation details.}

DeepSeek-V3 is used as the default backbone for heuristic generation and reflective refinement. Performance is evaluated using the optimality gap with respect to the best-known solution value, averaged across runs.

\subsection{Results}

\textbf{RQ1 [Effectiveness].} 

\paragraph{Online bin packing.}

Table~\ref{tab:binpacking} shows that \textsc{ReVEL} generally outperforms both handcrafted heuristics and prior LLM-based optimization methods across most BPP settings. The largest gains appear under tighter capacities and longer streams, where long-horizon search becomes increasingly difficult.

For example, under capacity $C=100$ and stream length $1$k, \textsc{ReVEL} reduces the optimality gap from $5.32\%$ (First Fit) and $3.78\%$ (ReEvo) to $2.34\%$. At $C=300$ and length $10$k, \textsc{ReVEL} further achieves the best overall result with a gap of only $0.16\%$. These results suggest that reflective refinement is effective in challenging long-horizon online optimization regimes, where sustained adaptation and progressive heuristic improvement are critical for maintaining search quality.

\begin{table}[htbp]
\captionsetup{width=\columnwidth}
\caption{Online bin packing results. Fraction of excess bins to lower bound (lower is better).}
\label{tab:binpacking}
\small
\setlength{\tabcolsep}{3pt}
\renewcommand{\arraystretch}{1.12}

\centering
\begin{tabularx}{\columnwidth}{@{}cc*{5}{>{\centering\arraybackslash}X}@{}}
\toprule
\textbf{Capacity} & \textbf{Size} & \textbf{First Fit} & \textbf{Best Fit} & \textbf{EoH} & \textbf{ReEvo} & \textbf{ReVEL (ours)} \\
\midrule
\multirow{3}{*}{100}
& 1k  & 5.32\% & 4.87\% & 3.03\% & 3.78\% & \textbf{2.34\%} \\
& 5k  & 4.40\% & 4.08\% & 2.15\% & \textbf{0.80\%} & 1.13\% \\
& 10k & 4.44\% & 4.09\% & \textbf{0.33\%} & \textbf{0.33\%} & 0.59\% \\
\midrule
\multirow{3}{*}{300}
& 1k  & 1.34\% & 1.19\% & 0.60\% & 1.04\% & \textbf{0.30\%} \\
& 5k  & 0.93\% & 0.84\% & 0.63\% & 0.27\% & \textbf{0.24\%} \\
& 10k & 0.92\% & 0.86\% & 0.58\% & 0.19\% & \textbf{0.16\%} \\
\midrule
\multirow{3}{*}{500}
& 1k  & 0.25\% & 0.25\% & 0.25\% & 0.25\% & \textbf{0.25\%} \\
& 5k  & 0.50\% & 0.50\% & 0.50\% & 0.50\% & \textbf{0.50\%} \\
& 10k & 0.50\% & 0.47\% & 0.47\% & 0.47\% & \textbf{0.45\%} \\
\bottomrule
\end{tabularx}
\vspace{-0.3cm}
\end{table}

\paragraph{Traveling salesperson problem.}

Table~\ref{tab:tsp-bench} reports TSP results across instance sizes ranging from 10 to 200 nodes. \textsc{ReVEL} achieves the best performance on most scales and substantially improves over EoH on larger instances. In particular, on TSP200, \textsc{ReVEL} reduces the optimality gap from $15.58\%$ to $11.46\%$. Although ReEvo remains competitive on selected settings such as TSP100, the performance gap generally widens as problem complexity increases, suggesting that iterative reflective refinement improves scalability in high-dimensional heuristic search spaces.

\begin{table}[htbp]
\captionsetup{width=\columnwidth}
\caption{Traveling Salesman Problem (TSP) results. Optimality gaps (\%) relative to OR-Tools solutions.}\label{tab:tsp-bench}
\small
\setlength{\tabcolsep}{4pt}
\renewcommand{\arraystretch}{1.15}

\centering
\begin{tabularx}{\columnwidth}{@{}l*{5}{>{\centering\arraybackslash}X}@{}}
\toprule
\textbf{Heuristic}
 & \textbf{TSP10 (\%)} 
 & \textbf{TSP20 (\%)} 
 & \textbf{TSP50 (\%)} 
 & \textbf{TSP100 (\%)} 
 & \textbf{TSP200 (\%)} \\
\midrule
EoH      & 3.52 & 9.33 & 10.24 & 11.39 & 15.58 \\
ReEvo    & 4.22 & 6.74 & 11.63 & \textbf{11.01} & 15.58 \\
\midrule
ReVEL (ours) & \textbf{2.11} & \textbf{6.74} & \textbf{9.20} & 12.64 & \textbf{11.46} \\
\bottomrule
\end{tabularx}

\end{table}

Across both benchmarks, the results indicate that group-aware reflective refinement improves search robustness across optimization problems with substantially different combinatorial structures.


\begin{table}[H]
\centering
\caption{Online Bin Packing: Method Across Models}
\label{tab:tablechange}
\footnotesize
\setlength{\tabcolsep}{4pt}

\begin{tabular*}{\columnwidth}{l @{\extracolsep{\fill}} l c}
\toprule
\textbf{Heuristic} & \textbf{Model} & \textbf{Gap(\%)} \\
\midrule
ReVEL & deepseek-v3-0324      & 1.13 \\
ReVEL & kimi-k2-instruct     & 1.95 \\
ReVEL & qwen3-coder-480b-35a  & 1.41 \\
ReVEL & glm-4.5              & 1.20 \\
\bottomrule
\end{tabular*}
\end{table}


\paragraph{Effect of LLM backbone.}

We instantiate \textsc{ReVEL} using four diverse LLM backbones, including DeepSeek-V3, Kimi-K2, Qwen3-Coder, and GLM-4.5, spanning both code-specialized and general-purpose instruction-tuned models. As shown in Table~\ref{tab:tablechange}, \textsc{ReVEL} consistently achieves low final optimality gaps across all backbones, ranging from $1.13\%$ to $1.95\%$. Despite substantial differences in model architecture, training objectives, and coding capabilities, performance remains consistently competitive. This robustness suggests that \textsc{ReVEL}'s effectiveness is primarily attributed to its structured reflective evolutionary process rather than dependence on a particular LLM. More broadly, these results indicate that behavior-aware grouping and multi-turn reflective refinement constitute a backbone-agnostic optimization mechanism, allowing the framework to improve heuristic evolution across diverse models reliably.

\textbf{RQ2 [Heuristic Evolution].} 

\paragraph{Effect of grouping strategies.}

We evaluate how different grouping strategies influence heuristic refinement on TSP50 under five settings: no grouping, random grouping, pairwise grouping, without LLM refinement, and the proposed behavior-aware grouping. Let $G_t$ denote the population optimality gap at iteration $t$. We define the improvement as $
\Delta_t = G_{t-1} - G_t,
$ measure refinement consistency as the fraction of iterations yielding improvement.

\begin{table}[htbp]
\centering
\caption{Performance of Grouping Strategies}
\label{tab:grouping_effect}
\footnotesize
\setlength{\tabcolsep}{4pt}
\begin{tabular}{lcccc}
\toprule
\multirow{2}{*}{Setting} & \multirow{2}{*}{Gap (\%)} & \multicolumn{2}{c}{$\Delta_t$} & \multirow{2}{*}{Cons. (\%)} \\
\cmidrule(lr){3-4}
& & Mean & Var. & \\
\midrule
Random Grouping         & 15.97 & 0.86 & 1.79 & 45 \\
Without Grouping        & 17.60 & 0.72 & 1.48 & 40 \\
Pair-wise Grouping      & 15.55 & 0.92 & 1.67 & 45 \\
w/o LLM Refinement  & 11.80 & 1.26 & 2.05 & 55 \\
\textbf{Our Method}     & \textbf{9.20} & \textbf{1.61} & \textbf{2.84} & \textbf{65} \\
\bottomrule
\end{tabular}
\end{table}

Table~\ref{tab:grouping_effect} shows that naive grouping provides only limited gains: random and pair-wise grouping reach similar gaps of $15.97\%$ and $15.55\%$, compared with $17.60\%$ without grouping. Thus, simple grouping heuristics or fixed-pair comparisons do not reliably produce informative reflective context. Behavior-aware grouping without LLM refinement already reduces the gap to $11.80\%$, confirming that meaningful organization of heuristics is critical. The full \textsc{ReVEL} further improves the gap to $9.20\%$, raises mean improvement from $1.26$ to $1.61$, and increases consistency from $55\%$ to $65\%$. These results show that robust refinement requires both behavior-aware organization and semantic consolidation: grouping exposes meaningful behavioral structure, while LLM refinement converts it into more effective and consistently improving heuristic updates.

\begin{figure*}[htbp]
    \centering
    \includegraphics[width=\textwidth]{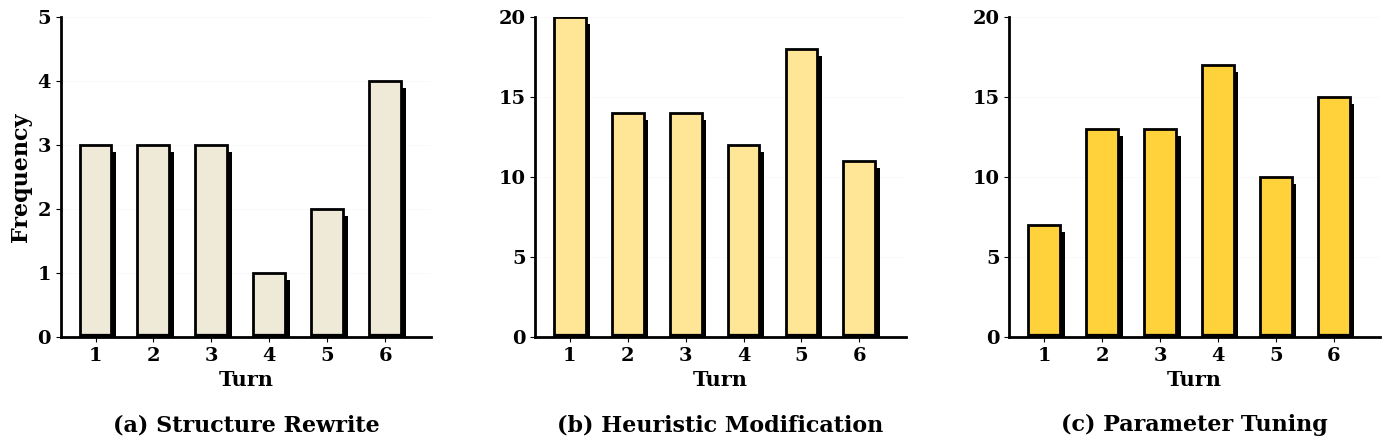}
    \vspace{-0.6cm}
    \caption{Distribution of refinement operation types across optimization iterations.}
    \label{fig:freq}
\end{figure*}

\textbf{Evolution of Reflective Refinement Strategies.}

Beyond final optimization performance, we analyze the internal refinement behavior of \textsc{ReVEL} throughout evolutionary search. Each reasoning trace is categorized into three refinement modes: \emph{structural rewrites}, \emph{heuristic modifications}, and \emph{parameter tuning}, using expert-defined criteria with LLM-assisted annotation and expert verification.

Table~\ref{tab:reasoning_breakdown} shows that refinement is dominated by heuristic modifications (\textbf{89}) and parameter tuning (\textbf{75}). At the same time, structural rewrites are relatively rare (\textbf{16}), indicating that \textsc{ReVEL} primarily improves existing heuristic families instead of repeatedly replacing them. Figure~\ref{fig:freq} further reveals a clear coarse-to-fine refinement trajectory: structural rewrites dominate early iterations, whereas localized refinement increasingly prevails as optimization progresses. Notably, structural rewrites continue to appear intermittently, preserving exploration even in later stages. These findings provide mechanistic evidence that \textsc{ReVEL} adaptively adjusts its refinement granularity according to accumulated multi-turn feedback, leading to more robust and sustained heuristic evolution.

\textbf{RQ3. [Ablation Study].} 

\begin{table}[htbp]
\centering
\caption{Ablation study on TSP50.}
\label{tab:ablation_study}
\small
\setlength{\tabcolsep}{3pt}
\begin{tabular*}{\columnwidth}{@{\extracolsep{\fill}} l r @{}}
\toprule
\textbf{Setting} & \textbf{Gap (\%)} \\
\midrule
Single-turn Refinement & 17.18 \\
Best-of-$N$ Sampling & 16.68 \\
Random Grouping & 15.97 \\
Without Grouping & 17.60 \\
\midrule
\textbf{\textsc{ReVEL} (Full)} & \textbf{9.20} \\
\bottomrule
\end{tabular*}
\end{table}
Table~\ref{tab} isolates the contribution of \textsc{ReVEL}'s two core mechanisms: behavior-aware grouping and sustained multi-turn refinement. Removing either component sharply degrades performance, increasing the TSP50 gap from {9.20\%} to {17.18--17.60\%}. Best-of-$N$ sampling also remains substantially worse ({16.68\%}), showing that additional candidate generation alone cannot reproduce the gains of iterative reflection.

Random grouping provides only a modest improvement over no grouping ({15.97\%} vs.\ {17.60\%}), indicating that grouping itself is insufficient; the reflective context must preserve meaningful behavioral structure. Taken together, the ablations rule out both increased sampling and naive population partitioning as explanations for the improvement. Instead, \textsc{ReVEL}'s advantage arises from the interaction between behavior-aware grouping, which organizes informative population-level context, and multi-turn reflection, which progressively exploits that context across the optimization trajectory.



\paragraph{Effect of similarity balancing.}Table~\ref{tab:coe} studies the effect of similarity balancing coefficient $\alpha$. Performance is maximized at $\alpha=0.5$, whereas 

\begin{wraptable}{r}{4.5cm}
\vspace{-1.2em}
\centering
\small
\caption{Effect of similarity balancing coefficient $\alpha$ on TSP50.}
\label{tab:coe}

\begin{tabular*}{3.6cm}{@{\extracolsep{\fill}}lc}
\toprule
\textbf{$\alpha$} & \textbf{Gap (\%)} \\
\midrule
0.2 & 15.79 \\
0.4 & 14.06 \\
0.5 & \textbf{9.20} \\
0.6 & 15.97 \\
0.8 & 12.50 \\
\bottomrule
\end{tabular*}

\vspace{-1.0em}
\end{wraptable}


 emphasizing either behavioral or structural similarity alone results in substantial degradation. This suggests that effective reflective grouping requires alignment between optimization behavior and program structure, enabling more coherent reflective feedback and stronger optimization performance. \\
\textbf{RQ4. [Cost Efficiency].}

\hspace{-0.3cm}\textbf{Cost comparasion.} The average API cost under identical settings is reported in Table~\ref{tab:api_cost}. Although \textsc{ReVEL} incurs a modest increase in API cost (\$0.41 vs. \$0.25--\$0.33), it consistently achieves substantially better optimization performance. More importantly, these improvements are practically meaningful: on online BPP, they reduce up to 20 bins compared with the strongest baseline, directly lowering downstream operational cost. This demonstrates a favorable cost--quality trade-off, where a small increase in inference cost yields disproportionately larger optimization gains. To ensure fairness, we further extend EoH and ReEvo to match \textsc{ReVEL}'s API budget. Their performance remains largely unchanged, indicating that the performance gap cannot be explained by additional LLM calls alone. Instead, the benefit arises from how optimization feedback is accumulated and utilized throughout reflective evolution.


\begin{wraptable}{r}{5.0cm}
\vspace{-1.0em}
\centering
\small
\caption{Average API Consumption Costs per Run}
\label{tab:api_cost}

\begin{tabular*}{0.95\linewidth}{@{\extracolsep{\fill}}lc}
\toprule
\textbf{Method} & \textbf{API Cost} \\
\midrule
ReEvo & \$0.25 \\
EoH & \$0.33 \\
\textbf{ReVEL (Ours)} & \textbf{\$0.41} \\
\bottomrule
\end{tabular*}

\vspace{-0.8em}
\end{wraptable}

\begin{figure}[H]
    \centering
    \includegraphics[width=\columnwidth]{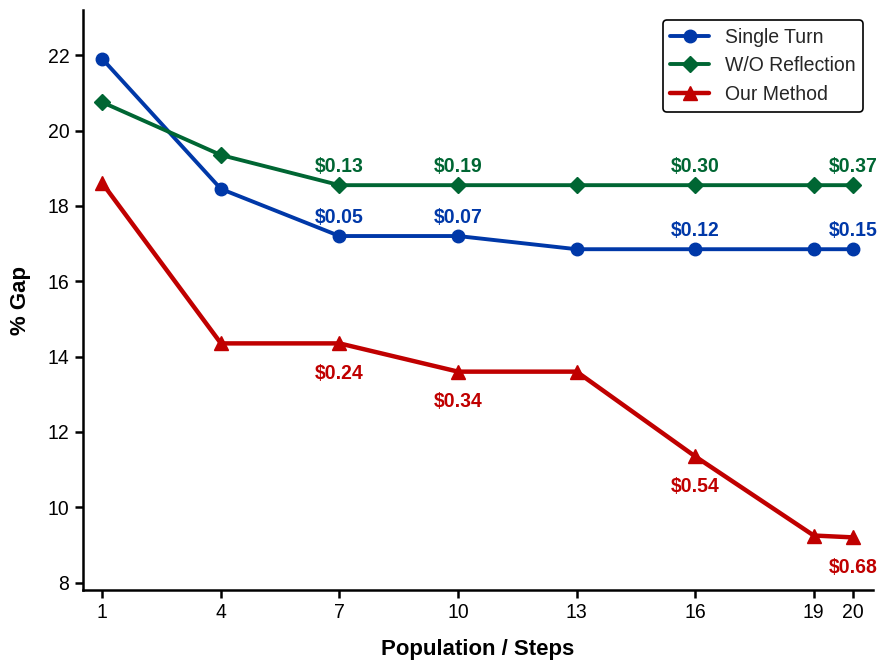}
    \caption{Cost--performance trade-off across optimization iterations.}
    \label{fig:gap1}
    \vspace{-0.5cm}
\end{figure}

\hspace{-0.3cm}\textbf{Cost-Performance Trajectory Analysis.} Figure~\ref{fig:gap1}

further analyzes optimization quality as API cost increases. Both the single-turn variant and the w/o reflection baseline improve rapidly during early iterations but plateau after approximately 10 optimization steps, converging to an optimality gap of around $18\%$. In contrast, \textsc{ReVEL} continues improving throughout the optimization trajectory, ultimately reducing the gap to nearly $9\%$.

This behavior suggests that the additional API budget is invested in progressively more informative reasoning rather than redundant heuristic generation. Without multi-turn reflection, extra LLM calls quickly become saturated because each refinement is conditioned only on local feedback. In contrast, \textsc{ReVEL} continuously accumulates optimization knowledge across iterations, allowing subsequent refinement decisions to become increasingly informative. Consequently, its performance continues improving long after competing methods have converged, demonstrating that the effectiveness of \textsc{ReVEL} stems from reasoning efficiency rather than inference budget alone.

\section{Conclusion}
\label{sec:conclusion}

We presented \textsc{ReVEL}, a reflective evolutionary framework for LLM-guided heuristic optimization. By combining behavior-aware grouping with multi-turn reflective refinement, \textsc{ReVEL} transforms heuristic evolution from repeated candidate generation into cumulative, feedback-driven optimization over evolving heuristic populations. Across diverse benchmarks and LLM backbones, \textsc{ReVEL} achieves strong and robust performance, while ablation and trajectory analyses show that its gains arise from structured population feedback and sustained reflective reasoning rather than increased sampling or inference budget alone. These findings highlight cumulative population-level reflection as a promising principle for more effective LLM-guided optimization.

\section{Limitations}

\noindent {\em \underline{Limitations}.}
While structured performance feedback enables informed multi-turn reflection, it does not guarantee uniformly beneficial refinements. In some cases, group-level feedback may obscure fine-grained behavioral differences, leading the LLM to propose ineffective or redundant heuristic modifications. Although such proposals are filtered by evolutionary selection, this can reduce sample efficiency in certain stages of the search.

\noindent {\em \underline{External Validity}.}
Although the evaluation does not cover the full diversity of combinatorial optimization settings, \textsc{ReVEL} was evaluated across several structurally distinct and widely studied classes of combinatorial optimization problems (COPs), including TSP, CVRP, online Bin Packing, and TSPLib benchmarks. These tasks span routing, packing, and large-scale benchmark optimization scenarios with varying search characteristics and solution structures. Furthermore, \textsc{ReVEL} was tested using multiple large language model backbones, including \texttt{deepseek-v3-0324}, \texttt{kimi-k2-instruct}, \texttt{qwen3-coder}, \texttt{480b-35a}, and \texttt{glm-4.5} (Table~\ref{tab:tablechange}), spanning both code-oriented and general-purpose instruction-tuned models. Nevertheless, optimization performance may still vary under alternative model architectures, prompting strategies, or domain-specialized LLMs.

\noindent {\em \underline{Internal Validity}.}
The evolutionary process involves inherent stochasticity arising from initialization, variation operators, and LLM sampling. To control variance, all experiments were conducted with fixed random seeds, repeated across multiple independent runs.

\noindent {\em \underline{Construct Validity}.}
LLM behavior is sensitive to prompt formulation, and variations in lexical framing or structural emphasis may affect reflective reasoning. To ensure consistency, all prompts followed fixed templates with explicit structural constraints, as detailed in the Appendix.

\section*{Acknowledgements}

This research is funded by Vietnam National Foundation for Science and Technology Development (NAFOSTED) under grant number 102.01-2025.69.

\bibliography{custom.bib}

\appendix
\newpage

\section{Appendix}

\section{Detail of benchmark dataset}
\noindent\textbf{Baseline setting.} To ensure a fair comparison, we align the setup based on the shared evolutionary paradigm of both methods by standardizing the population size and number of generations, which determine selection pressure and search depth. Concretely, we use comparable configurations (a population size of 10 and 20 evolutionary generations), ensuring similar evolutionary dynamics while retaining original mechanisms and default hyperparameters to avoid tuning bias; note that exact LLM call counts may differ due to implementation-specific factors, so we focus on aligning the high-level evolutionary structure.
\subsection{Traveling Salesman Problem (TSP)}

\subsubsection{Problem Definition}
The Traveling Salesman Problem (TSP) consists of finding the shortest possible tour that visits each city exactly once and returns to the starting city.  
Each instance is defined by a set of $N$ cities with two-dimensional coordinates
\[
\mathbf{x}_i = (x_i, y_i), \quad i = 1,\dots,N,
\]
where $x_i, y_i \in [0,1]$.  

The objective is to find a permutation $\pi$ of $\{1,\dots,N\}$ that minimizes the total tour length
\[
L(\pi) = \sum_{k=1}^{N-1} d(\pi_k, \pi_{k+1}) + d(\pi_N, \pi_1).
\]

\subsubsection{Evaluation Metrics}
The distance between cities $i$ and $j$ is the Euclidean distance $d(i,j) = \sqrt{(x_i - x_j)^2 + (y_i - y_j)^2}$. We evaluate solution quality using the \textbf{Optimality Gap} relative to the OR-Tools baseline ($L_{OR}$):
\[
\text{Gap (\%)} = \frac{L(\pi) - L_{OR}}{L_{OR}} \times 100\%
\]
\noindent where $L(\pi)$ represents the total tour length. We report the average gap across all instances; a lower gap indicates higher proximity to the best-known solution.

\subsubsection{Problem Sizes and Instances}

We consider multiple TSP problem sizes in order to evaluate scalability across small, medium, and large instances. 
Table~\ref{tab:tsp_sizes} summarizes all problem sizes used in our experiments. 

\begin{table}[H]
\centering
\caption{TSP problem sizes}
\label{tab:tsp_sizes}
\begin{tabular}{lc}
\toprule
Dataset & Number of Cities \\
\midrule
TSP10 & 10 \\
TSP20 & 20 \\
TSP50 & 50 \\
TSP100 & 100 \\
TSP200 & 200 \\
\bottomrule
\end{tabular}
\end{table}

\subsection{Online Bin Packing Problem}

\subsubsection{Problem Definition}
In the online one-dimensional bin packing problem, items arrive sequentially. Each item must be assigned immediately to a bin of $C$ capacity. If no existing bin has sufficient remaining capacity, a new bin is opened.

Given a sequence of item sizes $s_1, s_2, \dots, s_T$ with $s_i \in (0,1]$, the objective is to minimize the total number of bins used. Reordering or moving previously packed items is not allowed.

\subsubsection{Evaluation Metrics}

For each instance, items are processed strictly in their arrival order (Online Bin Packing). Each item is either assigned to an existing bin with sufficient remaining capacity or placed into a newly opened bin. Once an item is placed, it cannot be moved or reassigned.

Let $B$ denote the total number of bins used by an algorithm for a given instance $I$. We compare the algorithm's performance against the standard \textbf{volume lower bound}, defined as:
\[
LB(I) = \left\lceil \frac{\sum_{i=1}^{T} s_i}{C} \right\rceil
\]
where $s_i$ is the size of item $i$, $T$ is the sequence length, and $C$ is the bin capacity. 

The efficiency is measured using the \textbf{Fraction of Excess Bins}, expressed as a percentage:
\[
\resizebox{\columnwidth}{!}{$
\text{Fraction of excess bins}(I) (\%) = \left( \frac{B - LB(I)}{LB(I)} \right) \times 100\%
$}
\]
This metric quantifies the relative overhead compared to the theoretical minimum. A value of $0\%$ indicates an optimal solution (matching the lower bound), while higher percentages reflect lower packing efficiency. Final results are reported as the average percentage across all test instances.

\subsubsection{Problem Sizes and Instances}
To evaluate performance across different scales, we construct datasets with varying bin capacities and number of items. 
Table~\ref{tab:bpp_sizes} lists all configurations, where each dataset is defined by a pair $(T, C)$ of number of items $T$ and bin capacity $C$.

\begin{table}[H] 
\centering
\caption{Online Bin Packing problem sizes}
\label{tab:bpp_sizes}
\setlength{\tabcolsep}{4pt} 
\resizebox{\columnwidth}{!}{%
\begin{tabular}{ccc}
\toprule
Dataset & Number of Items ($T$) & Capacity ($C$) \\
\midrule
BPP-1K-100   & 1K  & 100 \\
BPP-1K-300   & 1K  & 300 \\
BPP-1K-500   & 1K  & 500 \\
BPP-5K-100   & 5K  & 100 \\
BPP-5K-300   & 5K  & 300 \\
BPP-5K-500   & 5K  & 500 \\
BPP-10K-100  & 10K & 100 \\
BPP-10K-300  & 10K & 300 \\
BPP-10K-500  & 10K & 500 \\
\bottomrule
\end{tabular}%
}
\end{table}

\section{Preliminary}

\subsection{Evolutionary Algorithms}

Formally, let $\mathcal{H}$ denote the space of all feasible heuristics. The goal of an Evolutionary Algorithm (EA) is to identify an optimal heuristic $h^* \in \mathcal{H}$ that maximizes a fitness function $\mathcal{F}: \mathcal{H} \to \mathbb{R}$:
\begin{equation}
    h^* = \operatorname*{arg\,max}_{h \in \mathcal{H}} \mathcal{F}(h)
\end{equation}
The algorithm maintains a population state at generation $t$, denoted as $\mathcal{P}^{(t)} = \{h_1^{(t)}, h_2^{(t)}, \dots, h_N^{(t)}\}$, where $N$ is the population size. The evolutionary process describes the transition $\mathcal{P}^{(t)} \to \mathcal{P}^{(t+1)}$ via a stochastic mapping $\mathcal{T}$:
\begin{equation}
    \mathcal{P}^{(t+1)} = \mathcal{T}(\mathcal{P}^{(t)}; \mathcal{S}, \mathcal{V})
\end{equation}
where the transition relies on two primary operators:

\begin{enumerate}
    \item \textbf{Fitness Evaluation:} Every individual $h_i^{(t)}$ is evaluated to obtain a quality score $s_i = \mathcal{F}(h_i^{(t)})$. This score typically drives the selection pressure.
    
    \item \textbf{Selection Operator ($\mathcal{S}$):} This operator selects a subset of parents $\mathcal{P}_{parents} \subset \mathcal{P}^{(t)}$ based on their fitness. A common formulation is tournament selection, where a parent is chosen via:
    \begin{equation}
        h_{parent} = \operatorname*{arg\,max}_{h \in \mathcal{T}_{sub}} \mathcal{F}(h)
    \end{equation}
    \begin{equation}
        \mathcal{T}_{sub} \sim \text{Uniform}(\mathcal{P}^{(t)}, k)
    \end{equation}
    where $\mathcal{T}_{sub}$ is a random subset of size $k$.

    \item \textbf{Variation Operator ($\mathcal{V}$):} In standard EAs, this involves Crossover ($\mathcal{C}: \mathcal{H} \times \mathcal{H} \to \mathcal{H}$) and Mutation ($\mathcal{M}: \mathcal{H} \to \mathcal{H}$). 
    
    In the context of \textbf{LLM-driven optimization}, the variation operator is re-parameterized as a conditional generation task using a Large Language Model $\pi_\theta$. Given a prompt template $I$ and parent heuristics, the offspring $h_{new}$ is sampled from the model's distribution:
    \begin{equation}
        h_{new} \sim \pi_\theta(\cdot \mid I, \mathcal{P}_{parents})
    \end{equation}
    Here, the LLM acts as a semantic mutation operator, leveraging its pre-trained knowledge to propose meaningful code transformations rather than random bit-flips.
\end{enumerate}

Finally, the next generation $\mathcal{P}^{(t+1)}$ is formed by selecting the best candidates from the union of the current population and the newly generated offspring (elitism), ensuring monotonic improvement in the best-found solution over time.

\subsection{From Pairwise to group-aware Estimation}

A central challenge in guiding LLM-based evolution is effectively estimating the relative quality of generated solutions to provide constructive feedback. Our methodology is directly motivated by the recent shift in preference optimization—specifically, the transition from pairwise comparisons to group-relative evaluations.

\textbf{Limitations of Pairwise Comparison (DPO):}
Direct Preference Optimization (DPO)~\cite{rafailov2023direct} aligns models by maximizing the likelihood margin between a winning sample $y_w$ and a losing sample $y_l$. While effective for general alignment, this pairwise paradigm is suboptimal for evolutionary search. Evaluating a population of size $N$ via pairwise comparisons restricts the model's view to local differences, ignoring the global context of how a solution stands against the entire population distribution.

\textbf{The group-aware Shift (GRPO):}
Group Relative Policy Optimization (GRPO)~\cite{shao2024deepseekmathpushinglimitsmathematical} addresses this by introducing a \textit{group-aware} perspective. Instead of isolating pairs, GRPO evaluates a candidate $o_i$ relative to a group of outputs $\{o_1, \dots, o_G\}$ generated from the same prompt, using the group mean as a dynamic baseline:
\begin{equation}
    A_i = \frac{f(o_i) - \text{mean}(\{f(o_j)\}_{j=1}^G)}{\text{std}(\{f(o_j)\}_{j=1}^G) + \epsilon}
\end{equation}
\textbf{Adoption in Our Approach:}
Inspired by this specific advancement, we adopt the \textit{grouping mechanism} of GRPO as the core principle for our reflection module. We hypothesize that for an LLM to effectively critique and improve heuristics, it must analyze solutions not in isolation or pairs, but as a cluster. 

\subsection{Prompt}

\begin{tcblisting}{
    colback=gray!5,
    colframe=gray!80,
    breakable,        
    listing only,
    listing options={
        basicstyle=\small\ttfamily, 
        breaklines=true,            
        breakatwhitespace=true,     
        columns=fullflexible
    },
    title=Online Bin Packing Prompt Formulation
}
TASK SUMMARY

You are an AI assistant whose job is to iteratively produce and refine Python heuristic implementations for the Bin Packing Online Problem.  
You will be given an existing heuristic (or helper functions). Use multi-turn reasoning: at each turn you must reflect, then either **explore** a new heuristic family or **exploit** (refine) the last submitted heuristic, and finally receive an observation/feedback from the environment.

- 

### FUNCTION CONTRACT (must be strictly respected)
- Language: Python only. Only standard library and numpy allowed (if already used by provided code).
- Required signature:
    def score(item, bins)
- Input arguments:
    - item: int # size of current item
    - bins : Numpy arrays # the rest capacities of feasible bins, which are larger than the item size.
- Return: scores (Numpy array)
- Correctness rules:
    - 'item' is of type int
    - 'bins' and 'scores' are both Numpy arrays.
\end{tcblisting}

\begin{tcblisting}{
    colback=gray!5,
    colframe=gray!80,
    breakable,        
    listing only,
    listing options={
        basicstyle=\small\ttfamily, 
        breaklines=true,            
        breakatwhitespace=true,     
        columns=fullflexible
    },
    title=Travelling Salesman Problem Prompt Formulation
}
TASK SUMMARY

You are an AI assistant whose job is to iteratively produce and refine Python heuristic implementations for the Travelling salesman problem.
Given a set of nodes with their coordinates, \
you need to find the shortest route that visits each node once and returns to the starting node. \
The task can be solved step-by-step by starting from the current node and iteratively choosing the next node. \
You will be given an existing heuristic, Let use multi-turn reasoning: at each turn you must reflect, then either **explore** a new heuristic family or **exploit** the last submitted heuristic, and finally receive an observation/feedback from the environment.

### FUNCTION CONTRACT (must be strictly respected)
- Language: Python only. Only standard library and numpy allowed (if already used by provided code).
- Required signature:
    def select_next_node(current_node, destination_node, unvisited_nodes, distance_matrix)
- Input arguments: This function should accept 4 inputs: 'current_node', 'destination_node', 'unvisited_nodes', 'distance_matrix'
- Output:  The function should return 1 output: 'next_node'
- Correctness rules: 'current_node', 'destination_node', 'next_node', and 'unvisited_nodes' are node IDs. 'distance_matrix' is the distance matrix of nodes. All are Numpy arrays.
Do not give additional explanations.
\end{tcblisting}

\begin{tcblisting}{
    colback=gray!5,
    colframe=gray!80,
    breakable,        
    listing only,
    listing options={
        basicstyle=\small\ttfamily, 
        breaklines=true,            
        breakatwhitespace=true,     
        columns=fullflexible
    },
    title=Prompt Template for the THINK Step
}
ALWAYS REMEMBER THAT, LOWER fitness score = BETTER solution.
First, based on the evaluation result from <observation> or GROUP REFLECTION, 
you should do some critical reasoning about the previous approach(s) ABOUT: 
+ its logical algorithm
+ its heuristic components/hyperparamters/features specifically. Then, think about the affect of these parameters/hyperparamters to the fitness score result  (in detailed).

Then, you can: 

1. Explore a totally new approach, to make some experiments to get information.
OR 
2. Focus on the behaviour of the heuristic features/components from the fitness result to tune them and get better result from the test evaluation.

You are ONLY allowed to do reasoning, NOT to generate code.
Note that, your reasoning should be very BRIEF but STILL critical and concise, focus on analyzing the heuristic components/features.
At the last of your response, there must be one of the tags <explore> or <exploit>, which indicate your decision.
\end{tcblisting}

\begin{tcblisting}{
    colback=gray!5,
    colframe=gray!80,
    breakable,        
    listing only,
    listing options={
        basicstyle=\small\ttfamily, 
        breaklines=true,            
        breakatwhitespace=true,     
        columns=fullflexible
    },
    title=Prompt Template for the Exploration Phase
}
Now, BASED solely on your REASONING, generate EXACTLY ONE solution for exploring.
Your output MUST be exactly the SAME as the following format:
<explore>
<algorithm>
# clear and complete algorithm description of the proposed heuristic.
</algorithm>
<code>
# the completely new Python function implementation for the algorithm in <algorithm> : `score(...)` (only code inside `<code>`).
</code>
</explore>
OUTPUT RULE:
Always output exactly one <explore> block containing both <algorithm> and <code>, nothing else.
\end{tcblisting}

\begin{tcblisting}{
    colback=gray!5,
    colframe=gray!80,
    breakable,        
    listing only,
    listing options={
        basicstyle=\small\ttfamily, 
        breaklines=true,            
        breakatwhitespace=true,     
        columns=fullflexible
    },
    title=Prompt Template for the Exploitation Phase
}
Now, BASED solely on your REASONING, generate EXACTLY ONE solution for exploiting.
Your output MUST be exactly the SAME as the following format:
<exploit>            
<algorithm>
# Clear algorithm description of the improvements you're making to the selected algorithm
</algorithm>
<code>
# Complete and concise Python function implementation with your refinements: `score(...)`
</code>
</exploit>
OUTPUT RULE:
Always output exactly one <exploit> block containing both <algorithm> and <code>, nothing else.
\end{tcblisting}

\begin{tcblisting}{
    colback=gray!5,
    colframe=gray!80,
    breakable,        
    listing only,
    listing options={
        basicstyle=\small\ttfamily, 
        breaklines=true,            
        breakatwhitespace=true,     
        columns=fullflexible
    },
    title=LLM Prompt Template for Heuristic Population Clustering
}
You are given a list of candidate solutions for a heuristic problem.  
Each candidate is a JSON object with fields:  
- "id": unique identifier  
- "code": the code itself  
- "score": numeric performance value  

Task: Group candidates that likely have the same behavior
1. Compare candidates by, code structure, and score values.
2. Group candidates into clusters of similar behavior, idea, or performance.
3. For each group, output:
   - Group label (integer ID).
   - List of member IDs.
   - Shared characteristics that define this group (algorithm style, coding pattern, performance range).
   - Distinctive differences within the group (if any).
4. If a candidate does not fit any existing group, assign it as its own group.
Also consider performance similarity:  
- Candidates with very different scores (e.g., difference > 0.1 on a [0–1] scale or > 10% relative difference) should not be grouped, even if code is similar.  
- Candidates with similar algorithmic ideas and close scores should be grouped together.  

Output format:  
Return ONLY a JSON object:

### example
{
  "groups": [
    {
      "id": 1,
      "members": ["1", "7", "9"],
      "shared_characteristics": "...",
      "differences": "...",
      "score_range": {"min": 0.72, "max": 0.75}
    },
    {
      "id": 2,
      "members": ["3", "4"],
      "shared_characteristics": "...",
      "differences": "...",
      "score_range": {"min": 0.55, "max": 0.58}
    },
    ...
  ]
}
###

CANDIDATE SOLUTIONS:\
\end{tcblisting}



\pgfplotsset{compat=1.17}
\pgfplotsset{
    mybarplotstyle/.style={
        ybar,
        bar width=0.4cm,        
        axis on top=false, 
        yticklabel style={
            font=\footnotesize,
            text width=1.5em,       
            align=right             
        },
        ymin=0,
        xmin=0.5, xmax=6.5,     
        xtick distance=1,       
        xticklabel style={font=\footnotesize},
        yticklabel style={font=\footnotesize},
        ylabel style={font=\footnotesize},
        xlabel style={font=\footnotesize},
        xtick pos=bottom,       
        ytick pos=left,         
        width=\linewidth,
        height=4.5cm,           

        axis background/.style={fill=grey!10}, 
        
        /pgf/bar/draw=black,
        /pgf/bar/fill=green!60!black, 
    }
}

\subsection{Some Code Examples}
\subsubsection*{Example 1: Baseline Implementation}

\begin{tcblisting}{
    title=Python Implementation,
    colback=gray!5,
    colframe=gray!80,
    breakable,
    listing only,
    listing options={
        language=Python,
        basicstyle=\footnotesize\ttfamily,
        keywordstyle=\color{blue!70!black},
        commentstyle=\color{gray!70!black},
        stringstyle=\color{orange!80!black},
        numbers=none,
        breaklines=true,
        columns=fullflexible,
        showstringspaces=false,
        tabsize=4
    },
    title=Example: Simulated Annealing Node Selection
}
import random
import numpy as np

def select_next_node(current_node, destination_node,
                     unvisited_nodes, distance_matrix):
    """A single simulated-annealing move to decide the next node."""
    if len(unvisited_nodes) == 0:
        return int(destination_node)

    distances = np.array([
        distance_matrix[current_node][node]
        for node in unvisited_nodes
    ])

    nearest_idx = int(distances.argmin())
    nearest_node = int(unvisited_nodes[nearest_idx])

    T = max(0.1, len(unvisited_nodes) / 100.0)

    prob_best = np.exp(-0.0 / T)

    if random.random() < 0.8:
        return nearest_node
    else:
        excess = distances - distances.min() + 1e-6
        probs = excess / excess.sum()
        chosen = random.choices(
            unvisited_nodes,
            weights=probs,
            k=1
        )[0]
        return int(chosen)
\end{tcblisting}

\subsubsection*{Example 2: Improved Implementation}

\begin{tcblisting}{
    colback=gray!5,
    colframe=gray!80,
    breakable,
    listing only,
    title=Good Example: Adaptive Simulated Annealing,
    listing options={
        language=Python,
        basicstyle=\footnotesize\ttfamily,
        keywordstyle=\color{blue!70!black},
        commentstyle=\color{gray!70!black},
        stringstyle=\color{orange!80!black},
        numbers=none,
        breaklines=true,
        columns=fullflexible,
        showstringspaces=false,
        tabsize=4
    }
}
import random
import numpy as np

def select_next_node(current_node, destination_node,
                     unvisited_nodes, distance_matrix):
    """Adaptive simulated annealing with dynamic temperature
    and exploitation parameters."""
    if len(unvisited_nodes) == 0:
        return int(destination_node)

    distances = np.array(
        [distance_matrix[current_node][node]
         for node in unvisited_nodes]
    )
    nearest_node = int(unvisited_nodes[np.argmin(distances)])

    remaining_nodes = len(unvisited_nodes)
    initial_temp = max(5.0, remaining_nodes * 0.5)
    decay_rate = max(10.0, remaining_nodes * 0.8)
    T = max(
        0.1,
        initial_temp * np.exp(-remaining_nodes / decay_rate)
    )

    base_exploit_prob = min(
        0.9,
        0.5 + 0.4 * (
            1 - remaining_nodes / len(distance_matrix)
        )
    )

    if random.random() < base_exploit_prob:
        return nearest_node

    for node, distance in zip(unvisited_nodes, distances):
        if node == nearest_node:
            continue

        delta_energy = distance - distances.min()

        if (delta_energy <= 0 or
            random.random() < np.exp(-delta_energy / T)):
            return int(node)

    return nearest_node
\end{tcblisting}

Example 1 uses fixed hyperparameters, making the search behavior insensitive to the remaining problem size.

Example 2 introduces adaptive temperature scheduling and a dynamic exploitation probability, enabling broader exploration in the early stage and increasingly greedy decisions as the search approaches completion.
\section{Additional Result}





\subsection{Reasoning Behavior}

\begin{figure}[H]
    \centering
    \includegraphics[width=1\linewidth]{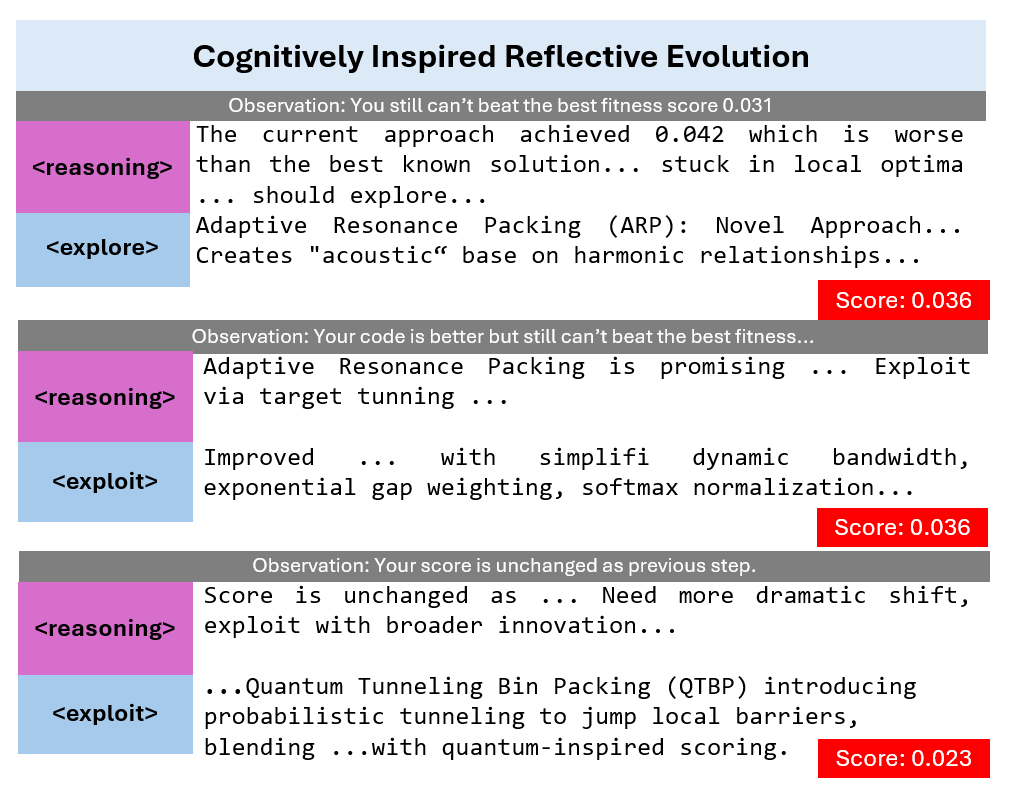}
    \caption{\textbf{Qualitative example of reflective reasoning in ReVEL:} Observations guide the LLM to alternate between exploration, exploitation, and innovation, resulting in progressive improvement of heuristic quality.}
    \label{fig:ReVEL_qual}
\end{figure}

As shown in Fig.~\ref{fig:ReVEL_qual}, the Reflective LLM-Guided Heuristic Evolution via Structured Performance Feedback process demonstrates how score dynamics and strategic adaptation interact to drive progress beyond local optima. At the outset, when the system observed a fitness of \(0.042\), considerably worse than the best-known score of \(0.031\), it recognized stagnation and initiated an exploratory shift. This led to the generation of Adaptive Resonance Packing (ARP), a structurally novel heuristic that improved the score to \(0.036\). Although this gain was modest, the introduction of ARP provided a fertile ground for subsequent refinements. 

Building on this foundation, the process transitioned into targeted exploitation. By tuning ARP through dynamic bandwidth control, exponential gap weighting, and softmax normalization, the system sought to consolidate and optimize the idea. While the score did not improve beyond \(0.036\), this stage illustrates the reflective nature of the method: rather than abandoning a promising approach, it strategically invested in fine-tuning, ensuring stability before pursuing further change.

When performance remained unchanged, the system’s observation mechanism signaled diminishing returns and prompted a more dramatic innovation. This shift produced Quantum Tunneling Bin Packing (QTBP), a probabilistic mechanism inspired by tunneling to bypass local barriers. Crucially, this step reduced the score to \(0.023\), surpassing the best-known baseline of \(0.031\). 

Our approach integrates exploration, exploitation, and reflective reasoning: exploration opens new directions, exploitation strengthens promising structures, and reflection decides when radical change is needed. Unlike naive retries, it uses performance signals to avoid stagnation and achieve breakthroughs beyond conventional search.

\subsection{Cost efficiency}
\begin{table}[htbp]
\centering
\caption{Average API Consumption Costs per Run}
\label{tab:api_cost}
\small
\begin{tabular}{l c}
\toprule
\textbf{Method} & \textbf{ API Cost} \\
\midrule
ReEvo & \$0.25 \\
EoH   & \$0.33 \\
\textbf{ReVEL (Ours)} & \textbf{\$0.41} \\
\bottomrule
\end{tabular}
\end{table}
The actual API costs under identical conditions are reported in Table~\ref{tab:api_cost}. Despite a modest increase in cost compared to the baselines, ReVEL consistently achieves superior performance.
Importantly, these gains are  practically meaningful: in BPP, they translate to up to 20 fewer bins, directly reducing operational cost. This reflects a stronger cost–quality trade-off, where a small additional budget yields substantially higher-quality solutions.

To further ensure fairness, we extend EoH and ReEvo to match ReVEL’s cost budget. The results remain consistent: both baselines plateau coverage early, as they rely on local, instance-level feedback that quickly becomes saturated around suboptimal solutions.

\subsection{TSPLib}
\paragraph{Evaluation on TSPLib Instances.}
\subsubsection{Evaluation on TSPLib Instances}

To further evaluate the practical effectiveness and robustness of ReVEL, we conducted additional experiments on a diverse set of TSPLib benchmark instances. Unlike the synthetic Euclidean datasets used in previous sections, TSPLib instances exhibit heterogeneous spatial structures, irregular node distributions, and varying scales. They are widely regarded as representative of practical TSP scenarios.

For each TSPLib instance, ReVEL was executed independently under the same configuration to evolve a heuristic specialized for that instance. Table~9 reports the percentage gap to the best-known TSPLib solutions and compares ReVEL with Nearest Neighbour and EoH.

\begin{table}[H]
\centering
\caption{Results on TSPLib instances. The gap (\%) to the best-known solution from TSPLib.
}
\label{tab:tsplib_comparison}
\setlength{\tabcolsep}{4pt}
\resizebox{\columnwidth}{!}{%
\begin{tabular}{cccc}
\toprule
Instance & Nearest Neighbour & EoH & Our \\
\midrule
Berlin52 & 19.06 & 17.68 & \textbf{6.24} \\
Bier127  & 15.27 & \textbf{8.04} & 9.33\\
Ch130    & 23.98 & \textbf{6.81}  & 10.21\\
D493     & 17.20 & \textbf{14.85} & 15.69\\
Eil51    & 19.72 & \textbf{3.19}  & 3.68\\
KroA100  & 26.19 & 11.06 & \textbf{6.21}\\
KroB100  & 31.68 & 9.84  & \textbf{9.65}\\
KroC100  & 26.88 & 19.31 & \textbf{5.88}\\
KroD100  & 26.56 & 13.21 & \textbf{10.01}\\
KroE100  & 25.01 & \textbf{11.79} & 13.46\\
Pr124    & 16.97 & 16.69 & \textbf{12.89}\\
Pr136    & 24.81 & 11.95 & \textbf{6.73} \\
Pr226    & 18.52 & 11.66 & \textbf{8.44}\\
Rat99    & 29.21 & 21.07 & \textbf{12.30}\\
St70     & 19.33 & 16.33 & \textbf{14.98}\\
Ts225    & 22.07 & 8.99  & \textbf{3.73}\\
U159     & 32.93 & 11.92 & \textbf{11.43}\\
Lin105   & 41.61 & \textbf{7.53}  & 21.37\\
\bottomrule
\end{tabular}%
}
\end{table}

Overall, ReVEL achieves the best results on 12 out of 18 instances and consistently improves upon the Nearest Neighbour baseline, confirming that the discovered heuristics substantially outperform classical constructive strategies. Compared with EoH, ReVEL shows competitive performance across most instances and provides substantial improvements on several structurally diverse problems, including Berlin52 (6.24\% vs.\ 17.68\%), KroC100 (5.88\% vs.\ 19.31\%), Rat99 (12.30\% vs.\ 21.07\%), and Ts225 (3.73\% vs.\ 8.99\%).

Notably, the performance advantage of ReVEL is particularly pronounced on instances with irregular spatial structures, where purely local constructive rules are insufficient. This suggests that the structured reflective feedback in ReVEL enables the LLM to discover heuristics that capture higher-level structural patterns rather than relying solely on greedy proximity decisions.

Across different instance sizes, ReVEL maintains stable performance without instance-specific parameter tuning, indicating that the proposed reflective evolution mechanism scales consistently across problem regimes.

These results demonstrate that ReVEL can reliably guide heuristic discovery across heterogeneous instance structures and problem scales, extending its effectiveness beyond synthetic benchmarks to realistic combinatorial optimization problems.

\begin{table}[H]
\centering
\caption{Hyperparameters Configuration}
\label{tab:hyperparameters}
\begin{tabular}{|l|c|}
\hline
\textbf{Parameter Name} & \textbf{Value} \\
\hline
\multicolumn{2}{|l|}{\cellcolor[gray]{0.9}\textbf{Evolution Loop Parameters}} \\
\hline
max\_turns & 6 \\
\hline
num\_candidates\_to\_initialize & 10 \\
\hline
epochs & 20 \\
\hline
top\_k & 10 \\
\hline
reminder\_probability & 0.3 \\
\hline
\multicolumn{2}{|l|}{\cellcolor[gray]{0.9}\textbf{Clustering \& Cross-over Parameters}} \\
\hline
num\_clusters & 3 \\
\hline
num\_elements & 4 \\
\hline
alpha & 0.5 \\
\hline
groups\_per\_crossover & 1 \\
\hline
\multicolumn{2}{|l|}{\cellcolor[gray]{0.9}\textbf{Evaluator Parameters}} \\
\hline
timeout\_seconds & 70 \\
\hline
\end{tabular}
\end{table}

\noindent
\textbf{Explanation of hyperparameters.}
As shown in Table ~\ref{tab:hyperparameters}, the hyperparameters are organized into three groups.
The \emph{Evolution Loop Parameters} control the overall evolutionary process, including the maximum number of interaction turns within each group (\texttt{max\_turns}), the initial population size (\texttt{num\_candidates\_to\_initialize}), the total number of evolutionary epochs (\texttt{epochs}), and the number of top-performing candidates preserved after each epoch (\texttt{top\_k}). 
The parameter \texttt{reminder\_probability} defines the probability of proactively injecting a reminder message to the LLM during the interaction loop, reinforcing awareness of the current best-fitness solution.

The \emph{Clustering \& Cross-over Parameters} determine how candidates are grouped and recombined.
Specifically, \texttt{num\_clusters} specifies the number of clusters formed from the population, \texttt{num\_elements} controls how many representative candidates are selected from each cluster for cross-over, and \texttt{alpha} balances semantic output similarity and code-level similarity during clustering.
The parameter \texttt{groups\_per\_crossover} enforces that each cross-over operation produces exactly one new interaction group.

Finally, the \emph{Evaluator Parameters} define execution constraints during evaluation.
The parameter \texttt{timeout\_seconds} limits the maximum runtime for evaluating each generated candidate, preventing non-terminating executions and ensuring stable evolutionary progress.



\subsection{Additional Baseline Results on CVRP}

The Capacitated Vehicle Routing Problem (CVRP) is a classical combinatorial optimization problem in which a fleet of vehicles with limited capacity must serve a set of customers with known demands while minimizing the total travel cost. Each vehicle starts and ends at a depot, and customer demands must not exceed vehicle capacity.


To assess the robustness of \textbf{ReVEL} beyond TSP and BPP, we conducted additional experiments on the \textit{CVRP} using the same experimental configuration as in the main paper. In this setting, heuristics were evolved independently for each problem size using an \textit{ACO-based backbone}, allowing ReVEL to adapt to different combinatorial structures and heuristic paradigms.

For a strong and meaningful comparison, we evaluated against \textbf{ReEvo}, a recent state-of-the-art LLM-guided heuristic evolution framework for this task. Since both methods operate under similar LLM-driven evolutionary settings, this comparison provides a controlled evaluation of the reflective evolution mechanism.

In table \ref{tab:cvrp-bench} ReVEL consistently outperforms ReEvo across all tested scales, with particularly large improvements on \textbf{CVRP20 (4.57\% vs.\ 13.70\%)} and clear gains on larger instances as well. These results demonstrate that the proposed reflective evolution framework is not limited to a specific problem type or heuristic structure, but can effectively guide heuristic discovery under different optimization paradigms.

Importantly, the CVRP experiments use the same default configuration without problem-specific tuning, indicating that ReVEL generalizes across combinatorial domains with minimal adjustment.

Overall, these additional results provide strong empirical evidence that ReVEL is a general LLM-guided heuristic evolution framework
, rather than a method specialized for a single problem class.

\begin{table}[H]
\centering
\caption{Baseline OR-Tools Results for the Capacitated Vehicle Routing Problem.}
\label{tab:cvrp-bench}
\setlength{\tabcolsep}{4pt}
\resizebox{\columnwidth}{!}{%
\begin{tabular}{cccc}
\toprule
\textbf{Heuristic}
 & \textbf{CVRP20 (\%)} 
 & \textbf{CVRP50 (\%)} 
 & \textbf{CVRP100 (\%)} \\
\midrule
ReEvo     & 13.70 & 19.44 & 18.06  \\
\rowcolor[gray]{0.93}
ReVEL (ours) & \textbf{4.57} & \textbf{17.68} & \textbf{14.88}  \\
\bottomrule
\end{tabular}%
}
\end{table}

\section{Cost-Performance Trade-off and Reflective Improvement Analysis}

\begin{figure*}[!t]
    \centering
    \includegraphics[width=0.96\textwidth]{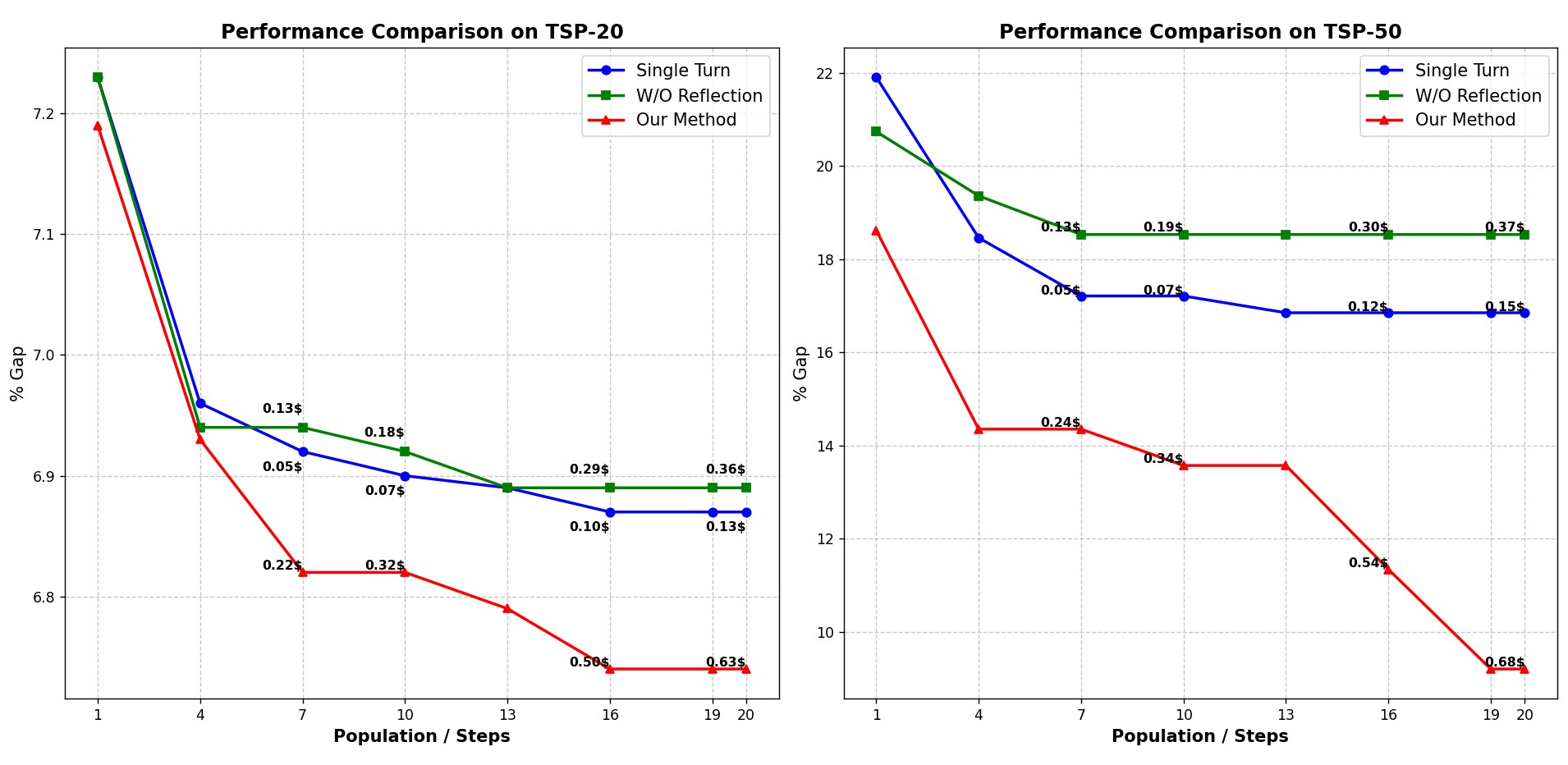}
    \captionof{figure}{Performance comparison of Single Turn, W/O Reflection, and our method on TSP-20 and TSP-50.}
    \label{fig:cost_tsp}
    \vspace{0.5em}
    \includegraphics[width=0.90\textwidth]{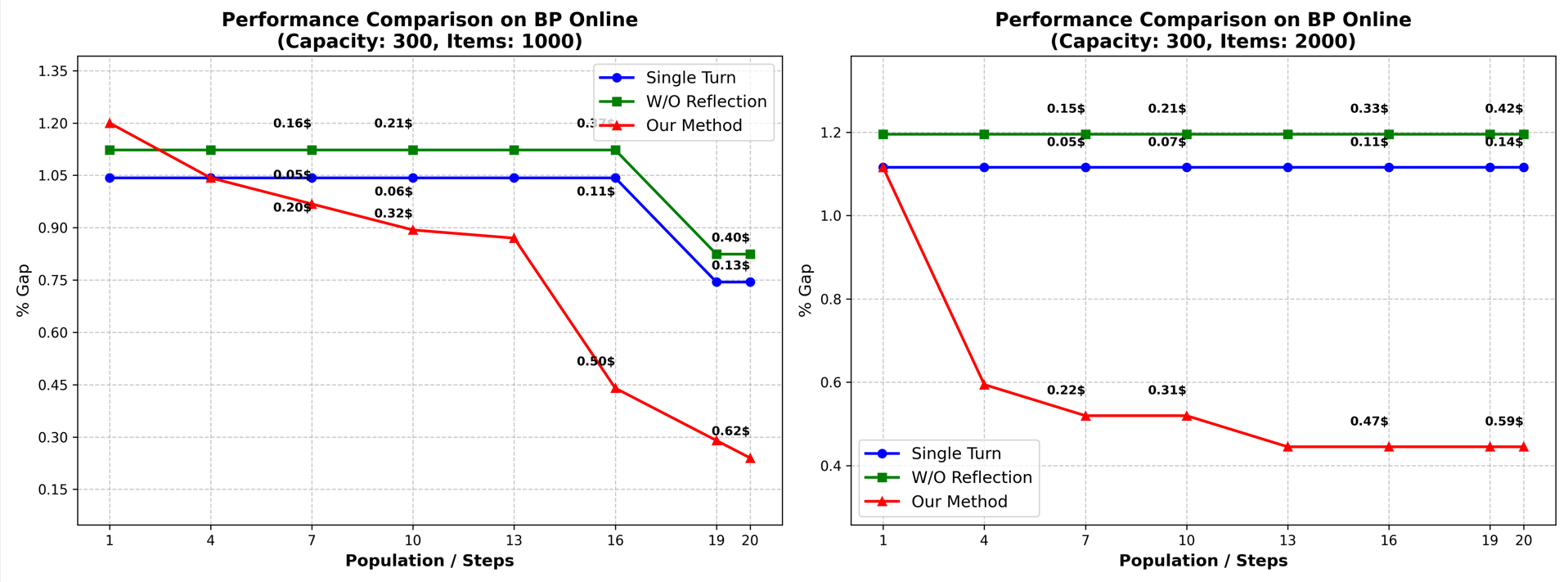}
    \captionof{figure}{Performance comparison on online bin packing with capacity 300 and 1000/2000 items.}
    \label{fig:bpo_1000_2000}
\end{figure*}

To better understand the efficiency and mechanism-level benefits of multi-turn reflection, we analyze both the computational cost and the performance trajectories of the generated heuristics. Figures~\ref{fig:cost_tsp} and~\ref{fig:bpo_1000_2000} report the optimization trajectories, along with the accumulated evaluation cost, for representative instances of the Traveling Salesman Problem (TSP20 and TSP50) and the Bin Packing Problem (BPP300-2000 items).

The results demonstrate a clear advantage in the cost-performance trade-off achieved by multi-turn reflection. On TSP50, ReVEL achieves a solution gap of approximately $9\%$ with a total cost of $0.68\$$. In contrast, single-turn reflection achieves only about $17\%$ gap at $0.15\$$, and the no-reflection baseline attains approximately $18\%$ gap at $0.37\$$. Similar behavior is observed on TSP20, where ReVEL consistently produces lower solution gaps while maintaining comparable computational cost.

The advantage of multi-turn reflection becomes even more pronounced in more challenging scenarios. On BPP instances with 300-2000 items, both single-turn reflection and the no-reflection baseline fail to generate any heuristic that improves upon the initial performance, even after the same number of evolutionary steps. In contrast, ReVEL continues to produce progressively improved heuristics under identical evaluation settings.

Although ReVEL requires a moderately higher computational cost, the improvement in solution quality is substantial. The additional cost mainly arises from two components:
\begin{itemize}
    \item \textbf{Pre-generation reflection}, which enables the language model to interpret structured performance feedback before heuristic synthesis;
    \item \textbf{Multi-turn refinement}, which allows heuristics to be iteratively improved rather than generated independently.
\end{itemize}

These mechanisms enable more effective exploration of the heuristic space and improve the utilization efficiency of the evaluation budget.

The trajectory analysis also provides evidence that multi-turn reflection leads to progressive improvement of heuristic quality. Both single-turn reflection and the no-reflection baseline exhibit rapid early improvement, followed by clear performance plateaus, even when given the same computational budget as ReVEL. This behavior suggests limited reasoning depth and restricted capability to refine heuristic logic.

In contrast, ReVEL demonstrates steady and sustained improvement throughout the reflective process. As structured feedback accumulates across reflection turns, the language model can refine heuristic strategies consistently and coherently. This results in a sequence of heuristics with progressively improved performance rather than a collection of largely independent candidates.

Overall, the results indicate that multi-turn reflection is not merely an additional computational expense but a mechanism that enables structured iterative improvement of heuristic programs. The proposed framework achieves a favorable cost-performance trade-off by producing substantially better heuristics with only a modest increase in computation while providing clear empirical evidence of progressive heuristic refinement.

\subsection{Prompt Robustness Analysis}

To evaluate the robustness of \textsc{ReVEL} to prompt formulation, we conduct an additional study on the TSP50 benchmark using three substantially different prompt variants: (1) the original prompt used throughout the paper, (2) a more detailed reasoning-oriented prompt, and (3) a concise prompt containing only the essential instructions. Each configuration is executed independently three times, and all reported results are averaged across runs.

The three prompt variants exhibit slightly different optimization trajectories during the early search stage. The reasoning-oriented prompt achieves the strongest initialization (17.9\% gap versus 18.2\% and 18.4\% for the original and concise prompts, respectively), indicating that richer instructions can modestly improve early exploration. However, all variants consistently converge to nearly identical final performance (approximately a 9.2\% gap) after reflective evolution.

These results suggest that \textsc{ReVEL} is largely insensitive to prompt formulation. Prompt design primarily influences the initialization of heuristic search, whereas iterative reflective evolution progressively accumulates optimization knowledge and corrects early differences across generations. Consequently, the final optimization quality is governed by the reflective evolutionary process rather than careful prompt engineering.

\subsection{Repeated-Run Robustness Analysis}
\label{sec:repeated_run_robustness}

LLM-guided evolutionary optimization is inherently stochastic due to random population initialization, LLM sampling, and evolutionary search. To evaluate the robustness of \textsc{ReVEL}, we repeat all main experiments over \textbf{10 independent runs} and report the mean and standard deviation of the final optimality gap. To enable comparison across benchmarks with different performance scales, we additionally report the coefficient of variation (CV), defined as $\mathrm{CV}=\frac{\sigma}{\mu},$ where $\mu$ and $\sigma$ denote the mean and standard deviation across runs.

\begin{table}[H]
\centering
\caption{Repeated-run robustness on Online Bin Packing over 10 independent runs. Results report mean optimality gap $\pm$ standard deviation and coefficient of variation (CV). Lower is better.}
\label{tab:bpp_repeated_runs}
\small
\setlength{\tabcolsep}{5pt}
\begin{tabular}{ccrr}
\toprule
\textbf{Capacity} & \textbf{Size} &
\textbf{\textsc{ReVEL} (\%)} & \textbf{CV} \\
\midrule
\multirow{3}{*}{$100$}
& $1$k  & $2.32 \pm 0.08$ & $0.0345$ \\
& $5$k  & $1.15 \pm 0.11$ & $0.0957$ \\
& $10$k & $0.53 \pm 0.16$ & $0.3019$ \\
\midrule
\multirow{3}{*}{$300$}
& $1$k  & $0.31 \pm 0.02$ & $0.0645$ \\
& $5$k  & $0.25 \pm 0.03$ & $0.1200$ \\
& $10$k & $0.18 \pm 0.06$ & $0.3333$ \\
\midrule
\multirow{3}{*}{$500$}
& $1$k  & $0.25 \pm 0.00$ & $0.0000$ \\
& $5$k  & $0.50 \pm 0.00$ & $0.0000$ \\
& $10$k & $0.46 \pm 0.01$ & $0.0217$ \\
\bottomrule
\end{tabular}
\end{table}

\begin{table}[t]
\centering
\caption{Repeated-run robustness on the Traveling Salesman Problem over 10 independent runs. Results report mean optimality gap $\pm$ standard deviation and coefficient of variation (CV). Lower is better.}
\label{tab:tsp_repeated_runs}
\small
\setlength{\tabcolsep}{7pt}
\begin{tabular}{lrr}
\toprule
\textbf{Benchmark} &
\textbf{\textsc{ReVEL} (\%)} & \textbf{CV} \\
\midrule
TSP10  & $2.33 \pm 0.61$  & $0.2618$ \\
TSP20  & $6.72 \pm 0.12$  & $0.0179$ \\
TSP50  & $9.80 \pm 0.79$  & $0.0806$ \\
TSP100 & $13.83 \pm 1.96$ & $0.1417$ \\
TSP200 & $12.54 \pm 2.03$ & $0.1619$ \\
\bottomrule
\end{tabular}
\end{table}
Tables~\ref{tab:bpp_repeated_runs} and~\ref{tab:tsp_repeated_runs} show that \textsc{ReVEL} exhibits stable final performance across independent runs despite stochastic population initialization, LLM sampling, and evolutionary search. Across Online BPP, the absolute standard deviation remains at most $0.16$ percentage points, indicating that repeated runs converge to closely comparable solution quality. Most configurations have a CV below $10\%$, while the $C=500$ settings exhibit near-zero variation because these regimes are already saturated and offer little remaining optimization headroom.

The larger CV values for BPP-10K-100 and BPP-10K-300 should be interpreted together with their absolute scale. Their mean gaps are only $0.53\%$ and $0.18\%$, respectively, so small absolute deviations of $0.16$ and $0.06$ produce comparatively large normalized ratios. Thus, the elevated CV reflects division by a near-zero mean rather than substantial instability in practical solution quality.

A similar pattern holds for TSP. From TSP20 to TSP200, CV remains between $1.79\%$ and $16.19\%$, with no systematic increase as the number of nodes grows. This is notable because larger TSP instances induce substantially broader heuristic search spaces and more opportunities for stochastic evolutionary trajectories to diverge. TSP10 exhibits the highest relative variation ($26.18\%$), but its absolute deviation remains only $0.61$ percentage points; the higher CV again partly reflects its small mean gap. Importantly, increased problem scale does not lead to uncontrolled variance: TSP100 and TSP200 retain moderate CV values despite their greater search complexity.

Overall, the repeated-run analysis indicates that stochasticity primarily affects the particular refinement trajectory taken by each run, rather than the quality regime ultimately reached. \textsc{ReVEL} repeatedly converges to similar final gaps across problem domains and scales, supporting that its reported gains are reproducible and not driven by favorable initialization or isolated LLM generations.

\end{document}